\documentclass{article}
\usepackage{amssymb}
\usepackage{amsthm}
\usepackage[margin=1in]{geometry}
\usepackage[utf8]{inputenc}
\usepackage{mathtools}

\usepackage[singlelinecheck=false]{caption}
\usepackage{graphicx}
\usepackage{pgfplots}
\usepackage{pst-solides3d}
\usepackage{sidecap}
\usepackage{wrapfig}

\graphicspath{{./graphics/}}
\pgfplotsset{compat=1.18}
\sidecaptionvpos{figure}{m}

\usepackage{centernot}
\usepackage{extpfeil}
\usepackage{extarrows}
\usepackage{faktor}
\usepackage{mathrsfs}
\usepackage{xcolor}
\usepackage{multirow}
\usepackage{enumitem}
\usepackage{booktabs}

\usepackage[section]{placeins}

\usepackage{hyperref}
\usepackage{cleveref}

\usepackage{tikz}
\usetikzlibrary{arrows}
\usetikzlibrary{automata}
\usetikzlibrary{calc}
\usetikzlibrary{cd}
\usetikzlibrary{decorations.markings}
\usetikzlibrary{decorations.pathmorphing}
\usetikzlibrary{shapes.geometric}

\tikzset{if/.code n args={3}{\pgfmathparse{#1}%
  \ifnum\pgfmathresult=1\pgfkeysalso{#2}\else\pgfkeysalso{#3}\fi}}
\tikzset{>=latex}
\tikzset{snake it/.style={decorate, decoration=snake}}
\tikzset{dbl/.style={>=stealth,
                     double,
                     double equal sign distance,
                     -implies,
                     shorten >=3pt,
                     shorten <=3pt}}          
\tikzstyle surjective=[postaction={decorate,decoration={markings, mark=at position .9 with {\arrow{latex}}}}]

\def\<{\langle}
\def\>{\rangle}

\DeclareMathOperator\arctanh{arctanh}

\makeatletter
\newcommand{\tpitchfork}{%
  \vbox{
    \baselineskip\z@skip
    \lineskip-.52ex
    \lineskiplimit\maxdimen
    \m@th
    \ialign{##\crcr\hidewidth\smash{$-$}\hidewidth\crcr$\pitchfork$\crcr}
  }%
}
\makeatother

\theoremstyle{definition}

\usepackage[minbibnames=2, backref=true]{biblatex} %Imports biblatex package
\usepackage{hyperref}
\usepackage{subcaption}

\usetikzlibrary{arrows.meta, positioning, calc, fit, backgrounds, decorations.pathreplacing}
\usepackage{bm}
\newcommand{\sign}{\operatorname{sign}}

\title{\textbf{Scaling Up Thermodynamic AI Models}}
\author{Andrew G. Moore \textit{et alia}}
\date{June 22, 2026}

\begin{document}

\maketitle

\begin{abstract}
Thermodynamic computing devices based on the Ising model show great promise for low-power AI inference and edge computing, but scalable methods for training large models for such hardware remain limited. Prior theory shows that the time-averaged behavior of high-temperature Gibbs-sampled Ising systems can implement feed-forward neural inference. We turn this theoretical correspondence into a scalable and purely backpropagation-based algorithm for training deep convolutional networks for thermodynamic inference on Ising machine hardware. Our image classification models achieve accuracies of 94.9\% on CIFAR-10 and 76.0\% on CIFAR-100 under binary Gibbs sampling. We then develop and experimentally validate a mathematical theory relating inference cost to accuracy and controlling autocorrelation times. Subsequently, we calculate asymptotic results showing that inference cost is bounded by a well-controlled tradeoff with performance and exhibit algorithms for computing optimal inference schedules. Finally, we discuss implications for hardware development and the future of high-temperature thermodynamic AI models.
\end{abstract}
    
% ======================================================================
\section{Introduction}
\label{sec:intro}
    
AI models aggregate many small computations into statistical inferences. However, running an AI model on a GPU, even a quantized one, entails exact error-free computation of every activation. This exactitude at a microscopic level, however, is not necessary for macroscopic performance: small errors in activations generally do not cause catastrophic performance loss. This is the basic principle behind weight quantization methods and dropout layers. On the other hand, keeping those computations error-free comes with an energy cost at the hardware level. In other words, keeping each computation exact during inference is wasteful. Thermodynamic computing hardware, including both stochastic silicon and other probabilistic substrates, promises to perform \textit{approximate} local computations at a tiny fraction of the energy expenditure. At the macro level, therefore, running a model on such hardware could perform computations far more efficiently without significant accuracy loss. Given the massive and ever-growing energy demands of AI inference, the need to find a more efficient solution is likely to become more pressing over time. 
    
Thermodynamic computing systems exploit thermal fluctuations, stochastic dynamics, Boltzmann distributions, or non-equilibrium processes rather than suppressing noise as an error source \cite{CCC2019ThermodynamicComputingWorkshop}. A canonical example is the Ising computer: any type of hardware which implements a system of binary spins whose probability distribution is governed by the Ising model. Ising computation has a long history with AI, going back to the 80s. The Hopfield network, a canonical example of the connection between statistical mechanics and machine learning, was explicitly based on the low-temperature Ising model. The Boltzmann machine, a positive-temperature cousin, is still relevant today. However, the crown jewels of modern AI inference, such as image classification and natural language processing, have largely remained outside their reach. 
    
Ising-based systems have always struggled with a lack of truly scalable and efficient training algorithms for general workloads. This has mostly confined Ising computation in practice to those situations in which a good training method does exist. In applications where a Boltzmann machine is suitable, contrastive divergence or equilibrium propagation provides a solution, but these methods do not scale efficiently to large benchmark problems \cite{Nguyen_2017}. Ising computers have also found wide application in combinatorial optimization, where weights can easily be derived analytically from the structure of the problem \cite{Lucas_2014}. However, the lack of a purely backpropagation-based approach has hampered the ability of the technology to tackle large-scale AI workloads: in short, the best methods we know for training models have not yet been applicable. Thermodynamic AI should not
require abandoning the modern deep-learning software stack. This paper aims to rectify that. We take an unorthodox approach to Ising computation, develop new backpropagation-based algorithms for training thermodynamic models, demonstrate our results on benchmark tasks, and provide theory to explain our results.

\subsection{Main Contributions}

The direct precursor to this work is \cite{moore2025correspondenceisingmachinesneural}, which establishes a correspondence theorem connecting shallow tanh neural networks to Ising systems by a global scaling parameter $\delta$. In theory, it promises that neural networks can be run directly on Ising machines. More specifically, by uniformly scaling down the interactions of the second layer interactions, the average behavior of the Ising system approximates the activations of a neural network with the same topology, weights, and tanh activation.  However, that result only shows that a trained shallow perceptron runs on an Ising machine, and only in a low-coupling and infinite-time limit. Furthermore, it does not address any concerns about mixing times or dynamics. In other words, the result remains a mathematical possibility, not a practical recipe. To make thermodynamic AI models a reality at practical workload scales, we need to be able to train deep models, control inference times, and ensure that the correspondence holds in a meaningful way at practical coupling scales. 

This paper is an algorithmic continuation and application of that work. We develop algorithms to train deep neural networks as a series of \textit{thermodynamic blocks}, where each block meets the criteria of the correspondence theorem (\Cref{sec:arch}). Thus, each block in the network can be implemented as a stochastic Ising machine, average-sampled at high temperature, with binary data passed between blocks. Furthermore, we introduce a method for ensuring that our models function at practical coupling scales, not only in a low-coupling limit (\Cref{sec:regularize}). Our training algorithm is based on pure backpropagation through regularization terms, ensuring that it scales with problem size. Finally, we develop a convergence theory and error rate estimates (\Cref{sec:convergence}), demonstrating that inference cost and mixing times remain under control, a necessary ingredient in using a stochastic system for practical applications. We develop asymptotic estimates and closed-form approximate formulae for the error rate as a function of inference time, then show, within a linearized red-black model, that the autocorrelation time goes to one in the low-coupling limit. This theory allows us to easily derive optimal inference-time schedules and be assured that the cost remains reasonable. 

On the implementation side, we demonstrate the power of our methods by training thermodynamic models on image classification benchmarks (\Cref{sec:results}). When running our models with thermodynamic inference, we achieve accuracies of 98.1\% on MNIST, 93.5\% on FashionMNIST, 94.9\% on CIFAR-10, and 76\% on CIFAR-100. We also demonstrate with selected results that there is a mild tradeoff between inference-time cost and performance: more extreme parameters and larger sweep counts can achieve better accuracy, but less extreme inference costs still produce good results. Analyzing the models, we show that, for our larger models, over $99.99\%$ of the FLOPs have been off-loaded to thermodynamic inference. In \Cref{sec:experiments}, we continue the experimental investigation by validating our theoretical models, visually characterizing model performance, conducting ablation experiments to show the impact of our methods, and explaining the procedure for deriving optimal inference schedules. 

Finally, in \Cref{sec:hardware}, we discuss how our results relate to the current thermodynamic hardware landscape. Any particular hardware system imposes unique constraints, especially on sparsity and quantization; therefore, we discuss how our algorithms can fit into that picture with an experiment. Furthermore, while the new approach of high-temperature sampling may be able to inject a new ingredient into the long-standing relationship between Ising computation and combinatorial optimization, ultimately thermodynamic hardware must be developed which is well-suited to the AI workloads of the future. 

Our methods focus primarily on convolutional networks. One reason is that they fit the theory---an unrolled conv with tanh activation is essentially a sparse MLP. Another reason is that hardware requires sparsity and locality, and convolutional architectures build that in naturally from the beginning. But also, we believe that convolutional networks are especially well suited to the workloads we need thermodynamic AI for. Low-power image processing would be extremely useful for small surveillance drones or security cameras. Applied to one-dimensional convolutions, the same approach could benefit implantable medical sensors, biometric wearables, and audio signal-processing devices. 

To achieve our results, we use an unorthodox model of the Ising system as a computation platform. Instead of annealing to zero temperature or sampling a single state from a Boltzmann distribution, as is traditional in Ising computation, we rely on empirical time-averages of the Boltzmann distribution, taken by averaging over many samples at a fixed temperature. By building the theory of high-temperature computation into a full-scale application, we show that there is merit to this new approach: high-temperature average sampling is a natural fit for neural inference, and first-class support for this sampling technique should be a priority in hardware development. 

\subsection{Relation to Previous Work}

This work is also related to neural training methods based on equilibrium or
bidirectional dynamics, including equilibrium-propagation-style approaches such
as \cite{E_P_2025}. Those methods demonstrate that deep networks can be trained
to tolerate iterative bidirectional relaxation. Our setting is different in two
important respects. First, the communicating activations are stochastic binary
spins rather than continuous states. Second, we deliberately avoid making the
main training procedure depend on long sampling-based gradient estimators. The
Gibbs chain supplies targets for regularization, but the gradient path remains
short, smooth, and compatible with standard backpropagation.

The closest benchmark comparison is \cite{Niazi_2024}, which demonstrates
image classification using Gibbs-sampled Ising systems and provides an
important baseline for the field. Our approach differs by using a more
conventional deep-learning training pipeline, by targeting deeper
convolutional architectures, and by emphasizing the sweep-count tradeoff needed
for larger models. The comparison is not merely one of accuracy: the goal is to
show that thermodynamic inference can be made to look like a scalable neural
software stack rather than a collection of separately engineered small Ising
classifiers.

On the hardware side, the architecture proposed in
\cite{jelinčič2025efficientprobabilistichardwarearchitecture} is especially
relevant. That work articulates a system-level picture in which probabilistic
sampling units communicate through binary states and are supplemented by a
small amount of conventional computation at the input and output. Our model is
closely aligned with that picture: it uses a classical encoder and decoder
around a sequence of Ising-compatible thermodynamic blocks. The contribution
here is the training and inference recipe needed to make such blocks perform
deep neural computation at nontrivial scale.

Finally, this work should be distinguished from the broader Ising-machine
literature on combinatorial optimization. Quantum annealers, CMOS annealers,
digital annealers, coherent Ising machines, simulated bifurcation machines, and
probabilistic-bit systems all provide important evidence that Ising-like
substrates can be engineered and scaled. However, most of that literature
targets optimization: construct an energy function and search for a low-energy
state. The present paper uses Ising dynamics differently. We do not anneal to a
ground state. We use finite-temperature Gibbs dynamics to estimate Boltzmann
averages, and we train a neural architecture so that the signs of those
averages implement feed-forward inference.

\section{Architecture}
\label{sec:arch}

\begin{figure}
    \centering
    \includegraphics[width=0.8\linewidth]{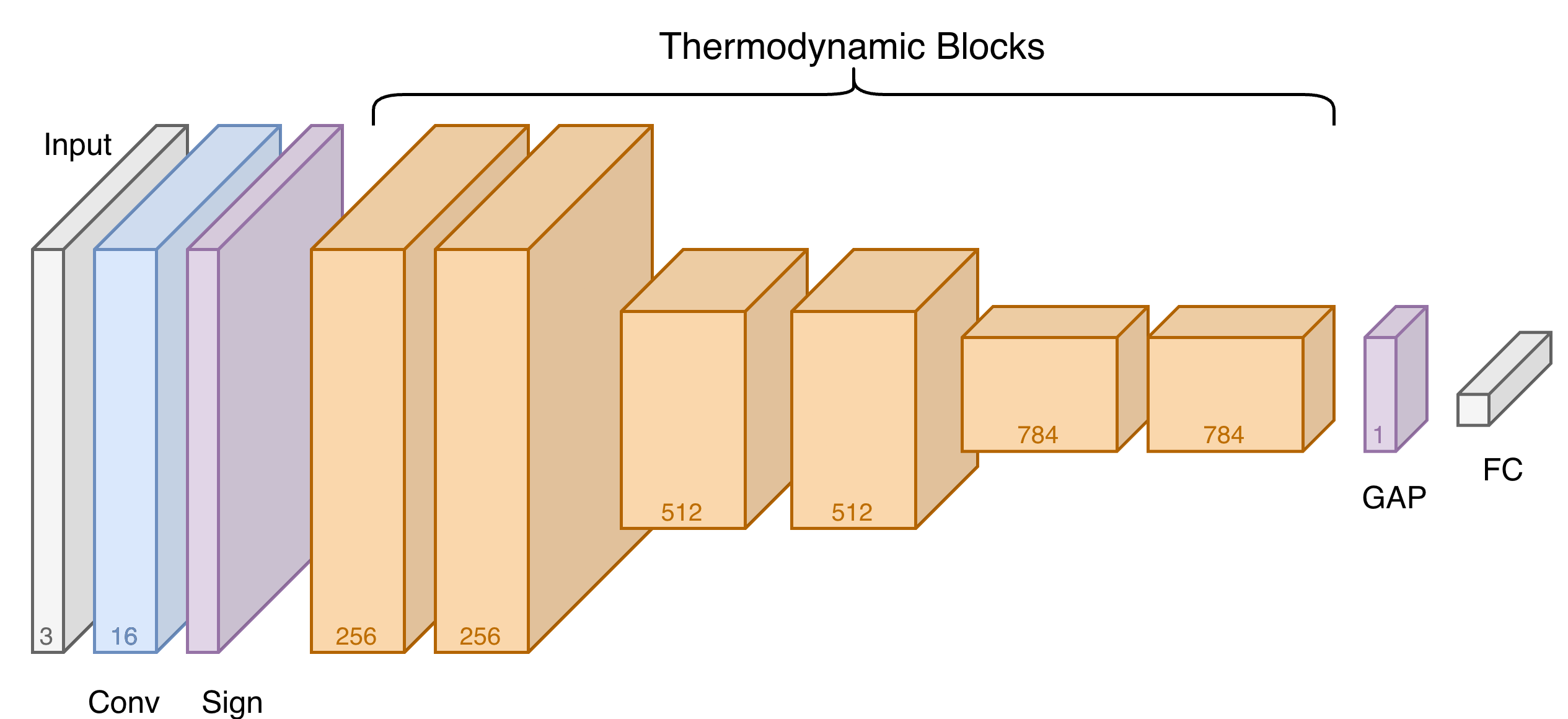}
    \caption{Our CIFAR-10 classifier. The encoder is a single classical convolution with sign activation. This is followed by six thermodynamic blocks. The classical decoder is a global average pool followed by a fully-connected linear map to the output layer. Each thermodynamic block contains three convolutional layers, as depicted in \Cref{fig:blockarch}.}
    \label{fig:architecture}
\end{figure}

Our model is loosely based on the Wide Residual Network (WRN) \cite{zagoruyko2017wideresidualnetworks}, a classic and well-tested convolutional architecture. It consists of an encoder, a sequence of thermodynamic blocks, and a decoder. The thermodynamic blocks account for between $99\%$ and over $99.99\%$ of the FLOPs.\footnote{To speak precisely, they \textit{replace} what would have been FLOPs in classical inference.} Each thermodynamic block is composed of two tanh-activated convolutional layers with a skip-convolution, followed by sign-activation to ensure that data flowing between blocks is binary. The model can be run as a normal feed-forward model on a GPU. On the other hand, each thermodynamic block can also be loaded onto an Ising machine and run with Gibbs sampling: the neurons are implemented as spins, and the weights and biases are simply reinterpreted as the coupling strengths and local magnetic fields, respectively. We envision the hardware implementation of the model as follows: classical GPU computation or dedicated digital ASICs perform the lightweight encoder and decoder. Each thermodynamic block is implemented by an Ising machine chip. Binary information is extracted from a block and passed to the next block as input. This setup is closely aligned with the overall hardware architecture proposed in \cite{jelinčič2025efficientprobabilistichardwarearchitecture}, although we use a completely different software model and training system. A visual representation of our model architecture is shown in \Cref{fig:architecture}.

\subsection{The Thermodynamic Block in Feed-Forward Mode}
\label{sec:feedforwardblock}

When run on a GPU in feed-forward mode, a thermodynamic block is a simple composition of convolutions, affine transforms, and hyperbolic tangent activations, with a final output passed through sign activation. There is nothing intrinsically probabilistic or energy-based about the block when run in this way. However, its architecture is carefully chosen so that it is capable of mapping directly onto probabilistic and energy-based Ising hardware. Each block is parameterized by three 
convolution operators $K_1, K_2, K_3$, one input affine map
\[
A_{\mathrm{in}}(x) = a_{\mathrm{in}} \odot x + d_{\mathrm{in}},
\]
one internal affine map
\[
A_{\mathrm{mid}}(u) = a_{\mathrm{mid}} \odot u + d_{\mathrm{mid}},
\]
and a hidden bias $b$. In our models, the convolutions $K_1$ and $K_2$ use a $3 \times 3$ kernel, while the skip-convolution $K_3$ uses a $1 \times 1$ kernel. Given a binary input tensor $x$, the sign-forward block computation of output binary tensor $y$ is
\begin{align}
x' &= A_{\mathrm{in}}(x),\\
z_1 &= K_1 x'  + b, \\
z_2 &= K_2 A_{\mathrm{mid}} \tanh(z_1) + K_3 x', \\
y &= \sign(z_2).
\end{align}
Thus the internal state of the block is continuous, but the quantity exported to the next block is
binary. The continuous internal states correspond approximately to thermal mean-field averages of the block when run on an Ising machine. The exported output must be binary to satisfy the requirements of inter-chip communication with Ising hardware, but this binarization also serves another purpose: clamping outputs to binary rounds away much of the error introduced by mapping the block into an Ising machine. 

\subsection{The Thermodynamic Block as an Ising Machine}

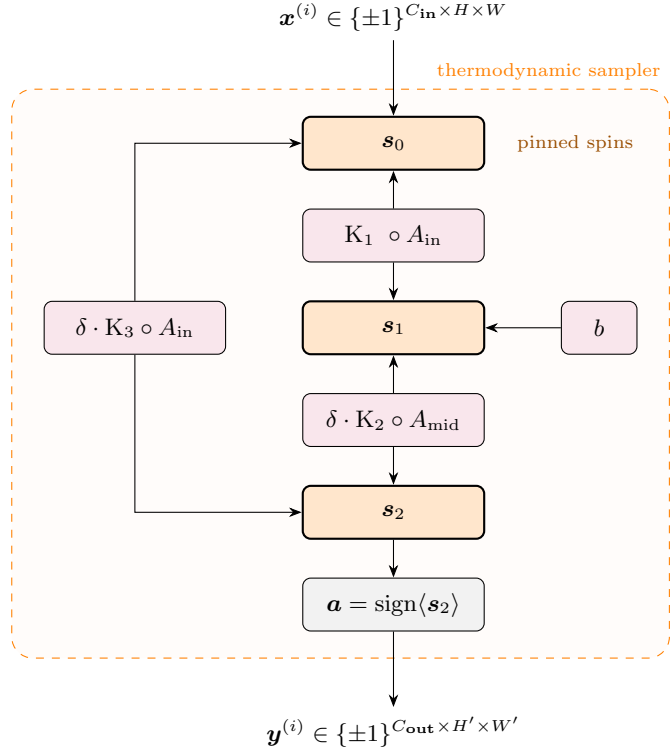
\begin{figure}
    \centering
    \begin{tikzpicture}[
>=Stealth,
node distance=1.1cm and 2.0cm,
op/.style={draw, rounded corners=3pt, minimum width=2.4cm, minimum height=0.7cm,
           fill=white!95!black, font=\small},
sign/.style={draw, rounded corners=3pt, minimum width=2.4cm, minimum height=0.7cm,
             fill=orange!20, font=\small, thick},
affine/.style={draw, rounded corners=3pt, minimum width=2.4cm, minimum height=0.7cm,
               fill=purple!10, font=\small},
binary/.style={font=\small\bfseries, text=black},
cont/.style={font=\small, text=orange!60!black},
every edge/.style={draw, ->, thick},
label/.style={font=\scriptsize, text=black!60},
]
% --- Input ---
\node[binary] (input) {$\bm{x}^{(i)} \in \{{\pm}1\}^{C_\text{in}\times H\times W}$};

\node[sign, below=1.0cm of input] (aff_in) {%
$\bm{s}_{0}$};
\node[cont, right=0.3cm of aff_in, font=\scriptsize] (sinplabel) {pinned spins};
% --- Conv1 ---
\node[affine, below=0.5cm of aff_in] (conv1) {$\text{K}_1\ \circ A_{\text{in}}$};
\node[sign, below = 0.5cm of conv1] (sign1) {$\bm{s}_1$};
\node[affine, minimum width=1cm, right=1.0cm of sign1] (bias1) {$b$};

% --- Conv2 ---
\node[affine, below=0.5cm of sign1] (conv2) {$\delta \cdot \text{K}_2\circ A_\text{mid}$};
% --- Conv3 (skip) ---
\node[affine, left=1cm of sign1] (conv3) {$\delta \cdot \text{K}_3\circ A_{\text{in}}$};
% --- z2 field ---
%\node[cont, right=0.3cm of sum, font=\scriptsize] (z2label) {$\bm{z}_2$};
% --- sign(z2) ---
\node[sign, below=0.5cm of conv2] (sign2) {$\bm{s}_2$};
\node[op, below=0.5cm of sign2] (avg) {$\bm{a} = \sign\langle \bm{s}_2\rangle$};
% --- Output ---
\node[binary, below=1.0cm of avg] (output) {$\bm{y}^{(i)} \in \{{\pm}1\}^{C_\text{out}\times H'\times W'}$};
% ===================== EDGES =====================
% Input -> Affine
\draw[->] (input) -- (aff_in);
% Affine -> Conv1
\draw[->] (conv1) -- (aff_in);
\draw[->] (conv1) -- (sign1);
\draw[->] (bias1) -- (sign1);
% Conv1 -> branch
%\draw[-] (conv1) -- (branch);
% Branch -> sign(z1)
%\draw[->] (branch) -| (sign1);
% Branch -> tanh
%\draw[->] (branch) -| (tanh1);
% tanh -> Conv2
\draw[->] (conv2) -- (sign1);
% Conv2 -> sum
\draw[->] (conv2) -- (sign2);
% Skip: aff_in -> Conv3 -> sum
\draw[->] (conv3) |- (sign2);
\draw[->] (conv3) |- (aff_in);
% sign2 -> output
\draw[->] (sign2) -- (avg);
\draw[->] (avg) -- (output);
% ===================== ANNOTATIONS =====================

% Dashed box around internal continuous region
\begin{scope}[on background layer]
\node[draw=orange!100, dashed, rounded corners=8pt, fill=orange!2,
      inner xsep=12pt, inner ysep=10pt,
      fit=(aff_in)(conv1)(sign1)(conv2)(conv3)(bias1)(sign2)(avg)] (contbox) {};
\node[font=\scriptsize, text=orange!100, anchor=south east] at (contbox.north east)
  {thermodynamic sampler};
\end{scope}
% s1 annotation
%\node[label, anchor=north, yshift=-2pt] at (sign1.south) {Ising spin (internal)};
% s2 annotation
%\node[label, anchor=east, xshift=-4pt] at (sign2.west) {Ising spin (output)};
% Ising field annotations
%\node[label, anchor=west, xshift=2pt] at (z2label.east) {local field};
%\node[label, anchor=west, xshift=2pt] at (z1label.east) {local field};
\end{tikzpicture}
    \caption{Architecture of a single thermodynamic block, implemented as an Ising system. It contains three layers of spins: input spins $s_0$, hidden spins $s_1$, and output spins $s_2$. The input layer $s_0$ is pinned in place. They are connected by bidirectional linear couplings, depicted in purple boxes. Couplings in an Ising system are bidirectional, but the labels indicate the forward coupling, i.e. moving downward in the diagram. The reverse couplings are the linear adjoint; for the exact relationship, see \Cref{eq:energy}. The hidden layer is also affected by the local magnetic fields $b$. Finally, an average-and-sign circuit $a$ accumulates time-averages of $s_2$ and outputs their signs.}
    \label{fig:blockarch}
\end{figure}

In \Cref{sec:feedforwardblock}, we describe the feed-forward operation of the thermodynamic block. In this section, we will describe how the parameters can be mapped directly to an Ising system. A critical extra ingredient is added to perform this mapping: a real-valued output coupling scale $\delta \in (0,1]$, which we call the \textit{temperature gradient parameter}. The parameter $\delta$ multiplies all couplings and biases applied to the output layer. The name `temperature gradient' comes from an analogy: modifying the temperature effectively scales all the couplings at once through changes in $\beta$. Therefore, scaling down all the couplings on one side of the system is analogous to increasing the temperature on that side. Thus, a smaller value of $\delta$ corresponds to a stronger `temperature gradient.' Smaller values of $\delta$ allow the interaction between the input layer and the hidden layer to be dominant over the interaction between the hidden layer and the output layer, encouraging forward information flow at the cost of increased noise. 

To perform the mapping, we begin by observing that the affine and convolutional operations compose to form a network of sparse linear couplings, where the activation sites are identified as Ising spins. The three layers of activation sites are interpreted as three layers of Ising spins: the input spins $s_0$, the hidden spins $s_1$, and the output spins $s_2$. Written in inner-product notation, the \textit{Hamiltonian} (energy function) of the unit is 
\begin{align}
\label{eq:energy}
    H(s_0,s_1,s_2) = -\langle s_1, K_1a_\text{in} s_0\rangle - \delta\langle s_2, K_3a_\text{in} s_0\rangle - \delta\langle s_2, K_2 a_\text{mid} s_1\rangle \\- \langle s_1,b+K_1 d_\text{in}\rangle - \delta \langle s_2, K_3 d_\text{in} + K_2 d_\text{mid}\rangle,
\end{align}
where the couplings are on the first line and the biases on the second line.
We assume that we are able to fix the input spins $s_0$, so that they do not evolve over time. This allows us to feed stable input data into the block. On the other hand, the values of the hidden spins $s_1$ and the output spins $s_2$ are allowed to evolve. We also fix the inverse temperature $\beta$ at 1, the natural value given the coupling scale set by neural-network training. The statistical mechanics of Ising systems ensures that the values of these layers will be distributed according to a Boltzmann distribution on the energy function $H$. Letting $\Sigma_\text{mid} = \{\pm 1\}^{C\times H\times W}$ and $\Sigma_\text{out} = \{\pm 1\}^{C\times H\times W}$ be the configuration spaces of the hidden and output layers, where $C$ is the number of channels and $H \times W$ is the image size, we can write the Boltzmann distribution as
\begin{align}
    \mathbb{P}[s_1,s_2|s_0] = \frac{1}{Z[s_0]} \exp(-H(s_0,s_1,s_2)), && Z[s_0] = \sum_{s_1 \in \Sigma_\text{mid}} \sum_{s_2 \in \Sigma_\text{out} } \exp(-H(s_0,s_1,s_2)).
\end{align}

In order to operate the block, we set the input spins $s_0$ to our input binary tensor $x$, then allow the system to fluctuate according to the Boltzmann distribution. We observe the long-term time average of the output spins, the conditional expectation $\langle s_2 | s_0\rangle$, according to this distribution. Finally, we extract the output by applying the sign function to this average, i.e. $y = \sign\langle s_2 | s_0\rangle$. The block architecture is diagrammed in \Cref{fig:blockarch}. The reason that it is designed in this way is that previous theory \cite{moore2025correspondenceisingmachinesneural} guarantees that as $\delta$ approaches $0$, $\langle s_1 | s_0\rangle$ approaches $\tanh(z_1)$ and $\sign\langle s_2 | s_0\rangle$ approaches $\sign(z_2)$. In other words, for a strong enough temperature gradient parameter, the thermodynamic computing system closely approximates the result of the block when operated in feed-forward mode.

It should be noted that this approach to using an Ising system for computation differs significantly from those typical in the existing literature. Typically, Ising systems are used in one of two ways. In \textit{annealing}, the temperature is initially high, then decreased slowly as the system fluctuates, bringing the system toward a low-energy state. After the anneal, the state of the system is observed, extracting a candidate low-energy state. On the other hand, in Boltzmann-style sampling, the temperature is fixed, but the object of interest is the probability distribution itself: the output is extracted by observing an output-spin configuration at a sampling time, thus sampling directly from the Boltzmann distribution. In our setup, we observe \textit{Boltzmann averages}, which can be thought of as averaging over many samples from a Boltzmann machine. This approach, while non-traditional, does naturally fit into the existing hardware paradigm, and is necessary to make the correspondence with feed-forward neural networks work \cite{moore2025correspondenceisingmachinesneural}.

%Finally, there is the temperature gradient parameter $\delta$, which is extremely important. We know that in the limit of $\delta \rightarrow 0$ and number of sweeps going to $\infty$, the Gibbs output (see \Cref{sec:gibbs}) equals the output of the network in STE mode (see \Cref{sec:ste}) \cite{moore2025correspondenceisingmachinesneural}. Our goal is to train an STE with suitable regularization such that this limit can be mostly achieved with moderate $\delta$ and sweep number. 

%The model is composed of a sequence of blocks, surrounded by a small encoder and decoder. The blocks, diagrammed in \Cref{fig:blockarch}, are our focus, as they will be run on the thermodynamic sampling units. Like a DTCA \cite{jelinčič2025efficientprobabilistichardwarearchitecture}, we conceive of the model as a sequence of thermodynamically sampled blocks, with binary data copied from one to the next. Each block consists of three layers of Ising spins: the input, the hidden, and the output. The layers are sparsely coupled. We represent the couplings as affine-modified convolution layers. 

\subsection{Gibbs Sampling}
\label{sec:gibbs}

So far, we have explained how the thermodynamic block works from the perspective of equilibrium thermodynamics. However, we have not yet specified its \textit{dynamics}. While any Markov chain whose stationary distribution is the Boltzmann distribution would work, for this paper we will focus on chromatic Gibbs sampling with two colors (also called red-black sampling), corresponding to the hidden and output layers. This is a natural approach because, with the input fixed, the dynamically changing component of the block is a sparse bipartite graph. In order to run the block, we approximate the Boltzmann average of $s_2$ using finitely many samples. We will define an update of both layers to be a single `sweep' of the sampler, and we will refer to the number of sweeps executed as $G$. 

Thus, the Gibbs sampler provides the core inference algorithm by probabilistically evolving the spins in an Ising block. We will now specify its operation precisely. Let $\mathcal{B}(p)$ be a variant of the Bernoulli distribution which takes $1$ with probability $p$ and $-1$ with probability $1-p$. When applied to a vector, it is sampled element-wise and independently. Let $\sigma$ be the sigmoid function, i.e. $\tanh$ rescaled to have the range $(0,1)$. Then, the Gibbs sampling algorithm performs sweeps of the form
\begin{align}
    s_1 &\leftarrow \mathcal{B}(\sigma(2(\delta a_\text{mid} \odot K_2^\top s_2 + K_1A_\text{in}x+b))),\\
    s_2 &\leftarrow \mathcal{B}(\sigma(2\delta(K_2A_\text{mid}s_1 + K_3A_\text{in}x))).
\end{align}
Outputs are extracted by time-averaging the values of $s_2$, then applying the sign function. There is also a burn-in period of $G_b$ sweeps in which the values are discarded. Usually, $G_b = \lfloor G/4 \rfloor$, though it is sometimes useful to use $G_b = 0$. In short, the output, referred to as the \textit{Gibbs consensus}, is computed as 
\begin{equation}
y = \sign\left(\frac{1}{G - G_b} \sum_{g=G_b+1}^{G} s_2^{(g)}\right).
\end{equation}
where $s_2^{(g)}$ refers to the state of $s_2$ after sweep $g$. There is no backpropagation through this procedure: it is used only for inference, and to generate teacher tensors for the Gibbs regularizer discussed in \Cref{sec:regularize}.

The values of $s_1$ and $s_2$ are initialized differently depending on what we are using the Gibbs sampler for. In a real device, and therefore in our real inference, we use random i.i.d. Rademacher initialization. However, when we use Gibbs sampling in the training regularizer, this extra ingredient of realism actually just wastes computation. Instead, it is advantageous to start as close as possible to equilibrium in order to minimize the value of $G$ necessary to get a strong signal of the true equilibrium. This can be done with feed-forward initialization, which is described in \Cref{sec:init}.

This setup introduces the core problem of inference hyperparameter selection: given a trained model, we must select $G$ and $\delta$ to maximize accuracy under Gibbs sampling while minimizing sweep cost. Generally speaking, higher $G$ gives better accuracy at any choice of $\delta$, while lower $\delta$ gives better long-term accuracy but requires higher values of $G$ to achieve it. This hyperparameter selection problem is critical to extracting good performance from our trained models; we will analyze this problem thoroughly and provide heuristics, modeling, and theory for hyperparameter selection in \Cref{sec:analysis} and \Cref{sec:experiments}. For now, the important qualitative point is this: while the theory in \cite{moore2025correspondenceisingmachinesneural} promises that as $\delta \rightarrow 0$ and $G \rightarrow \infty$ the block's outputs under Gibbs sampling will approach the block's outputs under feed-forward evaluation, it \textit{does not} make any promises about the actual values of $\delta$ and $G$ required. In fact, without specialized training, obtaining useful accuracy from the thermodynamic block would require extremely low $\delta$, and, as a consequence, impractically high $G$. Training the network to perform well at more moderate $\delta$ and reasonably low sweep count $G$ is the purpose of our core training algorithm, to which we now turn. 

\section{Gibbs-Regularization Training}
\label{sec:regularize}

The full training procedure has four phases. The purpose of the first three phases, which we refer to as pretraining, is to turn a standard residual network into a network which is capable of being loaded onto an Ising machine---in other words, we create the blocks described in \Cref{sec:feedforwardblock}. A detailed description of the pretraining pipeline can be found in \Cref{sec:pretraining}. The output of the pretraining is a trained Straight-Through Estimator (STE). The purpose of the
fourth phase, which is our core algorithmic contribution, is to reshape that network's weights so that its feed-forward outputs remain stable
under the noise and backward coupling of the Ising system. The goal is to create a network which not only performs well in feed-forward operation, but also produces good accuracy at reasonable cost under Gibbs sampling. While the raw STE is technically a valid Ising system, as proven by the correspondence theorem, the purpose of this training is to turn a theoretical solution into a practically useful solution. To achieve this, we continue to train the STE while introducing two new regularizers, the \textit{fixed-point loss} (\Cref{sec:fp_loss}) and the \textit{magnitude regularizer} (\Cref{sec:mag_reg}). Taken together, we call this process \textit{Gibbs regularization}.

The goal of Gibbs regularization is to reshape the weights so that the STE activations of each block become stable fixed points, or near-fixed points, of the corresponding Gibbs dynamics. To achieve this goal, we run a new cosine-annealed STE training schedule while introducing our new regularizers according to a gentle curriculum. The overall loss function is 
\begin{equation}
  \mathcal{L} = (1 - \alpha)\,\mathcal{L}_{\text{CE}} + \alpha\,\mathcal{L}_{\text{KD}}
    + \lambda_{\text{FP}}\,\mathcal{L}_{\text{FP}}
    + \lambda_{\text{mag}}\,\mathcal{L}_{\text{mag}}
  \label{eq:total_loss}
\end{equation}
The cross-entropy classification loss $\mathcal{L}_{\text{CE}}$ and the knowledge distillation loss $\mathcal{L}_{\text{KD}}$ are the same as in pure STE training. The STE path carries the classification signal.\footnote{As a variation, some of the classification pathway can be sent through the $y_\text{gibbs}$ instead of the STE classification. This can help the decoder adapt to the Gibbs-sampled states it will actually receive, but it makes the STE training more difficult. This also has the interesting effect of yielding Gibbs-sampled accuracies which outstrip feed-forward accuracies; this is not generally expected otherwise.} However, we have added two new loss terms, the Gibbs fixed-point loss $\mathcal{L}_{\text{FP}}$ and the magnitude regularizer $\mathcal{L}_{\text{mag}}$, which we will describe shortly. Three curriculum schedules prevent the Gibbs stability objective from destabilizing training:
\begin{enumerate}[nosep]
    \item \textbf{Backward coupling ramp:} We introduce a parameter $q$ which linearly increases from 0 to 1.0 over the first 10 epochs. This parameter modulates the backward coupling in the calculation of the fixed-point loss, allowing the model to slowly adjust to the influence of backward couplings.
    \item \textbf{Gibbs sweep ramp:} The number of Gibbs sweeps per block increases from 2 to $G_{\text{train}}$ over 10 epochs. Early epochs see only short chains, which differ less from the feed-forward initialization. Later epochs require stability under longer Gibbs chains.
    \item \textbf{FP loss weight ramp:} $\lambda_{\text{FP}}$ ramps from 0 to its target value over 3 epochs, preventing the fixed-point loss from dominating before the model has begun adapting.
\end{enumerate}
We also use gradient clipping at $5.0$. Data augmentation includes random crops, horizontal flips, AutoAugment, and Cutout-16, though the exact augmentation protocol should be dataset-specific. The batch size is set to 64. We have observed that larger batch sizes result in much worse training performance, suggesting that the natural stochasticity of small batch sizes may be important for proper adaptation. 

\subsection{The Fixed-Point Loss}
\label{sec:fp_loss}

The fixed-point loss is the mechanism by which the Gibbs equilibrium informs the gradient update.
It consists of two terms per block, measuring whether the sign-forward outputs are self-consistent
under backward coupling from the Gibbs consensus. We begin by sweeping each block for $G_\text{train}$ sweeps, then extracting the empirical average $a = \langle s_2\rangle_{G_\text{train}}$ and setting $y_\text{gibbs} = \sign a$. To reduce noise, the initial state for Gibbs sampling is usually computed using the feed-forward activations, as described in \Cref{sec:init}. The value $y_\text{gibbs}$ does not have any gradients, but it can be used as a target value to penalize the network for drifting far from its feed-forward state under Gibbs sampling. It gives the model an empirical target for the block's equilibrium behavior.

The total fixed-point loss for the block is
$\mathcal{L}_{\text{FP}} = \mathcal{L}_{\text{FP}}^{(1)} + \mathcal{L}_{\text{FP}}^{(2)}$. The two terms measure Gibbs consistency on the hidden and output layers, respectively, and are defined as follows.

The first part of the fixed-point loss is, roughly speaking, the error between the feed-forward hidden layer and a rough mean-field estimate for the hidden layer factoring in the backward coupling from an output layer set equal to $y_\text{gibbs}$. It checks whether
the sign-forward hidden state remains self-consistent once backward coupling from the Gibbs consensus is
introduced. We define it as
\begin{align}
    z_1 &= b + K_1A_\text{in}x,\\
    \hat{z}_1 &= q\delta \cdot a_{\text{mid}} \odot K_2^{\top}y_\text{gibbs},\\
  \mathcal{L}_{\text{FP}}^{(1)} &= \left\|
    \sign z_1 - \tanh(z_1 + \hat{z}_1)
  \right\|^2.
  \label{eq:fp_s1}
\end{align}
This formula may be somewhat puzzling to readers, who might expect $\|\tanh z_1 - \tanh(z_1 + q\delta a_\text{mid} K_2^\top a)\|^2$. However, this more obvious form does not work as well in practice. Using $\sign a$ instead of $a$ and $\sign z_1$ instead of $\tanh z_1$ produces much better results. Empirically, the signed target produces a stronger stability signal than the corresponding tanh target. We interpret this as an implicit margin-enlarging objective: the loss penalizes not only disagreement with the Gibbs consensus, but also small output fields that would be vulnerable to finite-sweep fluctuations. Additionally, the stronger loss signal improves training speed. We should also note that setting the target to be a sign rather than a tanh also implicitly helps the magnitude regularizer, as it tries to push the spin averages estimated by the feed-forward pre-activation closer to $\pm 1$. 

The second term propagates that coupled hidden estimate forward and checks whether the resulting output remains aligned with the original feed-forward output. We assume that the mean-field value of the hidden layer is in fact $\tanh(z_1 + \hat{z}_1)$ and propagate it to the output layer, checking it against the inference-mode STE forward output. We set
\begin{align}
z_2 &= K_3 A_\text{in}x + K_2 A_\text{mid}\tanh z_1,\\
\hat{z}_2 &= K_3A_\text{in}x + K_2 A_\text{mid} \tanh(z_1+\hat{z}_1),  \\
  \mathcal{L}_{\text{FP}}^{(2)} &= \left\|
    \sign z_2 - \tanh (\delta \hat{z}_2)
  \right\|^2.
  \label{eq:fp_s2}
\end{align}
Again, the reader may have expected $\|\tanh (\delta z_2) - \tanh (\delta \hat{z}_2)\|^2$, and again we have found that this more obvious form does not perform well in practice. Using $\sign z_2$ increases the loss, exaggerating it in the correct direction, and encouraging spin-averages further from zero. This extra magnitude regularization ends up being helpful in practice. 

Note that all the gradients flow through the tanh, as the $\sign z_1$ and $\sign z_2$ terms have no gradients. Note also that $y_\text{gibbs}$ has no gradient, so this design yields clean and low-variance gradients with a short backpropagation tree: the stochastic Gibbs chain provides the \emph{target} (what the equilibrium looks like), while only smooth, differentiable operations contribute to the gradient.

\subsection{Feed-Forward Spin Initialization}
\label{sec:init}

In the preceding section, we did not describe how the Markov chain is initialized. As mentioned in \Cref{sec:gibbs}, in the inference mode we use the hardware-realistic initialization of i.i.d. Rademacher. However, for generating the fixed-point loss, it is more computationally efficient to initialize as close as possible to the equilibrium state, since we can get reliable low-noise estimates of the long-term Gibbs consensus with far fewer sweeps. Therefore, when using the Gibbs sampler in training, we initialize the spins using the feed-forward activations of the block, setting
\begin{align}
    s_1^{(0)} &= \sign(b + K_1A_\text{in}x),\\
    s_2^{(0)} &= \sign(K_3 A_\text{in}x + K_2 A_\text{mid}\tanh(b + K_1A_\text{in}x)).
\end{align}
This gets the system close enough to equilibrium to provide a good training signal. In practice, with this initialization even $G_\text{train} = 10$ is sufficient to provide good stability information for training, with only diminishing returns up to $G_\text{train} = 40$. 

As a variation, we sometimes do use random Gibbs initialization, as opposed to feed-forward, and/or eliminate burn-in by setting $G_b = 0$. The former encourages convergence toward the desired state from a broader set of initial conditions, while the latter increases the sample size at the cost of some noise. These modifications may help in some cases, but generally come at the cost of increasing $G_\text{train}$ and thus slowing down training. 

\subsection{Training-Time Temperature Gradient}

The fixed-point loss described in \Cref{sec:fp_loss} takes the temperature gradient parameter $\delta$ as a hyperparameter. Naively, one might assume that we should train with the same value of $\delta$ with which we intend to perform inference. However, this is not the case. In general, we will perform very few sweeps during training (say, $G_\text{train}=10$), and many more sweeps during inference. We want to train with a $\delta$ that is high enough that low $G$ gives good information about the equilibrium state; low $G$ and low $\delta$ will produce a $y_\text{gibbs}$ which is essentially random noise. On the other hand, we will perform inference with much lower values of $\delta$ and higher values of $G$ to extract high classification performance. 

Additionally, we have observed that networks trained with a particular value of $\delta_\text{train}$ perform well on $\delta < \delta_\text{train}$, moderately well at $\delta \approx \delta_\text{train}$, and very poorly for $\delta > \delta_\text{train}$. It follows that if we want a network capable of flexibly operating across different regimes, we need a higher $\delta_\text{train}$. However, $\delta_\text{train}=1$ is suboptimal for training as well: $\delta_\text{train}$ needs to be low enough that the Gibbs equilibrium is similar to the Gibbs equilibrium in the low-$\delta$ inference regime. Through experimentation, we have determined that $\delta_\text{train}=0.7$ is usually a good compromise. 

\subsection{Magnitude Regularizer}
\label{sec:mag_reg}

Pre-activations which are too far from zero or too close to zero are both problematic. Pre-activations too close to zero require large numbers of sweeps for convergence (see \Cref{sec:convergence}). When small pre-activations from the input layer occur in the hidden layer, the spin becomes more susceptible to being changed by backward coupling from the output layer, which is also undesirable. Pre-activations which are too far from zero result in longer autocorrelation times, and thus also in more sweeps required for convergence. Additionally, large pre-activations saturate the gradient of tanh, resulting in slower learning. A bilateral penalty on pre-activation magnitudes prevents pathological field strengths. $\mathcal{L}_{\text{mag}}$ is calculated as the average over all spins of 
\begin{equation}
 \text{ReLU}\big(\tfrac{m}{2} - |z|\big)^2
  + 0.1 \cdot \text{ReLU}\big(|z| - 3m\big)^2,
  \label{eq:mag_reg}
\end{equation}
where $z$ is the feed-forward pre-activation of the spin and $m$ is the target magnitude, a training hyperparameter which we usually set to $\frac{3}{2}$.

\section{Results}
\label{sec:results}

In \Cref{tab:results} we present selected results for the basic classification datasets MNIST, FashionMNIST, CIFAR-10, and CIFAR-100. The table shows classification accuracy under thermodynamic inference for each dataset. We also show several variations to demonstrate how cost and accuracy can be traded off. 

\begin{table}[h]
\centering
\caption{Results on major image classification datasets. `Accuracy' refers to the model's classification performance under thermodynamic operation, i.e. with each thermodynamic block run according to the Gibbs sampling algorithm, while `FF Accuracy' refers to the model's classification accuracy in feed-forward mode. `Parameters' denotes the number of trainable parameters in the feed-forward STE model, while `\% Classical' denotes the percentage of FLOPs in the feed-forward model not offloaded to Ising systems.}
\label{tab:results}
\begin{tabular}{r|ccccccc}
\toprule
Dataset & Accuracy & Total Sweeps & $\delta$ & Blocks & Parameters & FF Accuracy & \% Classical\\
\midrule
MNIST & 98.1\% & 186 & 0.2 & 1 & 80k &  98.14\% & 0.8157\% \\
FMNIST & 93.5\% & 1.5k & 0.3 & 3 & 0.9M &  91.56\% & 0.0776\%\\
FMNIST & 93.0\% & 150 & 0.3 & 3 & 0.9M &  91.56\% & 0.0776\%\\
CIFAR-10 & 94.9\% & 30k & 0.05 & 6 & 31M &  95.16\% & 0.00880\%\\
CIFAR-10 & 92.1\% & 2k & 0.1 & 6 & 31M &  92.79\% & 0.00880\%\\
CIFAR-100 & 76.0\% & 25k & 0.05 & 8 & 42M & 77.04\% & 0.00455\%\\
CIFAR-100 & 71.7\% & 4.2k & 0.1 & 8 & 42M &  75.18\% & 0.00455\%\\ 
\bottomrule
\end{tabular}
\end{table}

Two things may jump out to the reader about this table: first of all, our better results often come simply from better feed-forward accuracy. This is a result of weakening $\lambda_\text{FP}$. This change increases feed-forward accuracy, but requires more sweeps and lower $\delta$ to get close to it. The other thing that might surprise the reader is the fact that FMNIST accuracy exceeds the feed-forward accuracy. This is due to the use of a slightly different version of the training algorithm using algorithm tricks described as variations in \Cref{sec:regularize}.

We include a measurement of ``\% Classical'' for each model. This refers to the number of classical FLOPs required for thermodynamic inference as a percentage of the FLOPs required to run the model in classical feed-forward mode. Even in the smallest model, over 99\% of the FLOPs have been off-loaded to the thermodynamic hardware, whereas for the larger models essentially all of the computation is thermodynamic. For our CIFAR-10 model, featured in most of our analytical experiments in \Cref{sec:experiments} and \Cref{sec:convergence}, 99.99\% of the computation is offloaded to the thermodynamic units. This shows that if thermodynamic hardware achieves significant power-consumption reductions, our model could take advantage of them.

\subsection{Parameter Counting}

The parameter numbers we listed in \Cref{tab:results} count the number of classical parameters in the feed-forward version of the model. This refers to the number of parameters that need to be trained. It is not the same as the number of spins required to implement the model using Ising chips, nor the number of couplings present on-chip. Implementing the thermodynamic block on an Ising machine \textit{unrolls} the convolutions into sparse linear connections and uses a spin for each channel of each pixel. The block's affine transforms are folded into the linear couplings and biases. The number of biases for each block is equal to the number of spins. Performing the accounting for our 31M-parameter CIFAR-10 model, we obtain the total spin and coupling counts shown in \Cref{tab:accounting}. Note that these counts refer to the model before any quantization or sparsification is applied. Sparsification techniques can significantly compress the number of spin-spin couplings required. We discuss this in more detail in \Cref{sec:sparsification}.
\begin{table}[h]
    \centering
    \begin{tabular}{c|cccccc|c}
        \toprule
         Block & 1 & 2 & 3 & 4 & 5 & 6 & Total\\
         \midrule
         Spins & 541k & 786k & 524k & 393k & 231k & 151k & 2.63M\\
         Couplings & 646M & 1.26B & 940M & 1.26B & 611M & 747M & 5.46B\\
         \bottomrule
    \end{tabular}
    \caption{Total number of spins and spin-spin couplings for each thermodynamic block in the 6-block CIFAR-10 classifier model, before any post-training sparsification techniques.}
    \label{tab:accounting}
\end{table}

\section{Theoretical Model}
\label{sec:convergence}

Previous theory \cite{moore2025correspondenceisingmachinesneural} proves that low enough $\delta$ and high enough $G$ will give good performance. In \Cref{sec:regularize}, we presented algorithms designed to improve performance at moderate $\delta$ and $G$. However, in order to practically guide experiments and inference, we need to know how performance actually scales with these two parameters: for a certain performance target, what is the scaling behavior of $G$ as $\delta\rightarrow 0$? And how does the error rate behave as sweeps progress? This is a critical practical question, as a thermodynamic computing system which takes forever to compute is not much use to anyone. Broadly speaking, it turns out that the error rate is a simple power law in the sweep number, that the critical prefactor is the density of pre-activations near zero, and that sweep cost is roughly proportional to the inverse square of $\delta$.

In the following sections, we will answer these questions rigorously and precisify these heuristic statements. We will precisely pin down the sources of error, provide approximate formulae for calculating how error relates to $\delta$ and $G$, and demonstrate that the mixing times of our thermodynamic blocks are controllable. We will build an analytical model that allows us to predict and control error rates, and exhibit experiments demonstrating its power. The key observation in the following analysis is that our systems are quite large and weakly coupled---therefore, simply ignoring spin-spin correlations and assuming spatial independence allows us to predict system-scale statistical quantities with a high degree of accuracy. The deviations from this assumption can be cleanly identified by comparing theory and practice; as it turns out, our theory is accurate precisely where it needs to be to find the optimal inference hyperparameters for high classification performance. 

In \Cref{sec:analysis}, we investigate and define the error sources in our model. In \Cref{sec:gaussian}, we provide an effective and simple model using Gaussian mixtures for the distribution of activations, which is the critical quantity in statistical error estimation. In \Cref{sec:error_rate}, we use this model to estimate the effect of choosing $G$ using numerical integration, then provide a simplified closed-form mathematical solution. Finally, in \Cref{sec:autocorrelation}, we show that we can keep the mixing times under control, tightening the validity of the rest of the theory. Experimental investigations applying this theory to the analysis and tuning of our models follow in \Cref{sec:experiments}. 

\subsection{Sources of Error}
\label{sec:analysis}

Before we can analyze the thermodynamic performance parameters, it is necessary to discuss conceptually the taxonomy of error sources and their precise definition. Thermodynamic operation introduces errors relative to feed-forward operation at each block. Therefore, understanding the etiology of these errors and keeping them under control is vital to obtaining good inference performance by intelligent selection of $G$ and $\delta$. 

As we have discussed, a thermodynamic block operating on an Ising machine \textit{approximates} its feed-forward STE output. In general, when our models are run in thermodynamic mode by running each block with time-averaged Gibbs sampling, the classification accuracy of the network approximates its feed-forward classification performance. The lower the value of $\delta$ and the higher the value of $G$, the closer the thermodynamically sampled accuracy comes to the feed-forward accuracy. In general, however, the feed-forward accuracy is an upper bound to the thermodynamic performance, seeing as errors introduced by Gibbs sampling are essentially random noise and therefore extremely unlikely to help classification accuracy rather than simply introduce entropy.\footnote{It should be noted that algorithmic variations such as introducing $y_\text{gibbs}$ path into the cross-entropy signal can violate this assumption. However, we will focus on the simpler case where the STE training only uses the feed-forward signal.} Examples of this phenomenon can be see in our main benchmark results, \Cref{tab:results}. We observe, indeed, that the thermodynamic accuracy is slightly lower than the feed-forward accuracy. 

There are three parts to the error analysis. At the level of the binary tensors passed between blocks, we need to understand the errors introduced by the choice of $G$, and the errors introduced by the choice of $\delta$. At the level of the model as a whole, we need to understand the statistical relationship between these error sources and the rate of misclassification.
The first two factors, which comprise the underlying sources of error, can be analyzed mathematically. The third factor, which tackles how the per-spin error rates translate to decreases classification accuracy, is fundamentally empirical, and will be measured and modeled in \Cref{sec:exp_thermal}, \Cref{sec:exp_stochastic_only}, and \Cref{sec:exp_adaptive}. For now, we turn to defining the underlying error sources precisely.

\subsubsection{Stochastic Error}
\label{sec:stochastic_error_definition}

Stochastic error is the error introduced by the finiteness of $G$. For fixed $\delta$, as $G \rightarrow \infty$, we know that $\langle s \rangle_{G,\delta} \rightarrow \mu_\delta$, where $\langle \cdot \rangle_{G,\delta}$ indicates empirical average after $G$ sweeps and $\mu_\delta$ indicates the true Boltzmann average of $s$. This is a well-known consequence of the theory of Markov chains. For finite $G$, there will generally be a discrepancy: $\langle s \rangle_{G,\delta} \neq\mu_\delta$. In the output layer, if the sign is not changed, the error is tolerated. If the sign is changed, an error is introduced into the block output. This source of error is measured by the \textit{stochastic error rate}: 
    \begin{align}
        R(G,\delta) = \frac{1}{|s_2|} \sum_{i=1}^{|s_2|} \bm{1}\{\sign \langle s_{2,i} \rangle_{G,\delta} \neq \sign \mu_{i,\delta}\}.
    \end{align}
    In other words, the stochastic error rate $R$ of a particular block is the fraction of spins in the output layer of that block whose finite time averages differ in sign from their true long-term averages. We will see in \Cref{sec:totalR} that 
    \begin{align}
        R(G,\delta) \sim \frac{C}{\delta \sqrt{G}},
    \end{align}
    where $C$ is a constant dependent only on the weights of the block and the input. Practically speaking, stochastic error is combated by increasing the sweep count $G$.

\subsubsection{Thermal Error}

Thermal error is the error introduced by the choice of $\delta$. By the correspondence theorem \cite{moore2025correspondenceisingmachinesneural}, we know that $\mu_\delta/\delta \rightarrow z_2$ as $\delta \rightarrow 0$, and therefore $y_\text{gibbs} = \sign(\mu_\delta) \rightarrow \sign(z_2) = y_\text{STE}$ whenever the corresponding component of $z_2$ is nonzero, where $y_\text{STE}$ is the output of the block in feed-forward mode. Due to the effect of backward couplings, the Boltzmann spin average of the output spins $s_2$, which we refer to as $\mu_\delta$, will differ from $\tanh(\delta z_2)$. If it does not differ in sign, the discrepancy is tolerated. If it does differ in sign, an error is introduced into the block output. Error introduced by the discrepancy at finite $\delta$ is measured by the \textit{thermal error rate}:
    \begin{align}
        E(\delta) = \frac{1}{|s_2|} \sum_{i=1}^{|s_2|} \bm{1}\{\sign \mu_{i,\delta} \neq \sign (z_2)_i\},
    \end{align}
    where $\mu_{i,\delta}$ is the Boltzmann average of spin $i$. 
    This measures the fraction of spins whose sign-outputs will not converge to their feed-forward values no matter how long the system is sampled. Thermal error is combated by lowering the temperature gradient $\delta$. Larger values of $\lambda_\text{FP}$ result in lower thermal error at any given value of $\delta$. Generally, for $\delta > \delta_\text{train}$, thermal error is high enough to cause accuracy collapse. 

\subsubsection{Overall Error}

Thermodynamic sampling introduces random error on a per-block basis in two ways, stochastic and thermal error. At the scale of the whole model, errors in the internal activations translate statistically into errors in the output. Because errors are introduced at each block, the total error rate between the thermodynamically operated model and the feed-forward operated model increases with depth as errors accumulate. Due to this accumulation, deeper models require a combination of increased per-block error tolerance, lower $\delta$, and higher sweep count: increasing tolerance lowers the accuracy penalty of errors, decreasing $\delta$ lowers thermal error, and increasing $G$ decreases stochastic error.

\subsection{Modeling the Output Spin Distribution}
\label{sec:gaussian}

\begin{figure}
    \centering
    \includegraphics[width=\linewidth]{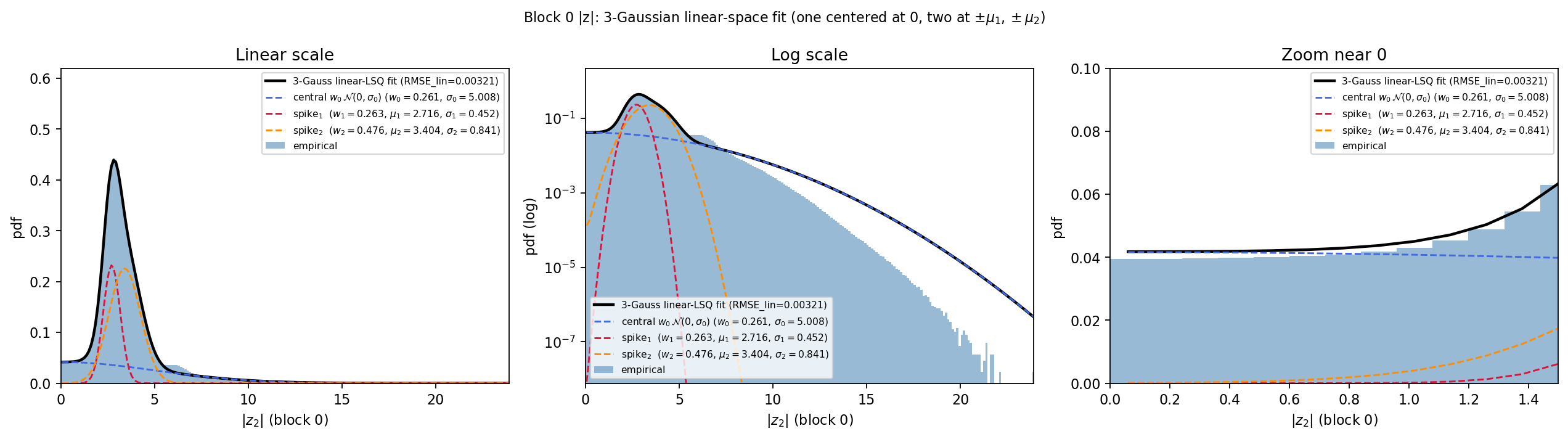}
    \caption{Empirical $|z|$ distribution and 3-component Gaussian mixture fit for the first block of our CIFAR-10 model.}
    \label{fig:mixture}
\end{figure}

In an Ising block, we have a very large number of output spins. We can calculate per-spin statistical quantities. However, the behavior of our systems is governed by a large-scale aggregation over these individual quantities. The aggregate behavior can differ qualitatively from the behavior of any individual component. For instance, we observe that each spin converges exponentially, but the overall error rate converges according to a power law. Given the very large number of spins, and the fact that the statistical behavior of each spin depends only on its long-term average activation, it is both justified and mathematically convenient to assume that a block contains an infinite number of spins governed by an absolutely continuous distribution. 

We begin by analyzing the distribution $\rho(|z|)$ of absolute pre-activations of output spins in the feed-forward model. Empirically, this can be measured by using a histogram. In \Cref{fig:mixture} we show a real example of a $|z|$-histogram. This shape is very typical. The peak is caused by the magnitude regularizer; almost all blocks trained with our method look like this. The next step is to choose an analytical approximation $\hat{\rho}$ of the distribution $\rho$ that fits well and is convenient to work with. A Gaussian mixture is effective, natural, and renders the analysis quite easy (empirically, $N=3$ mixture components is enough for a good fit). It is most important that $\hat{\rho}$ fits $\rho$ well near $z=0$, as this region turns out to govern the asymptotics of the stochastic error rate, but qualitative global fit is still desirable. To preserve behavior near $z=0$, when fitting the Gaussian mixture we constrain one of the spikes to be centered on $z=0$ and require that $\hat{\rho}(0)= \rho(0)$. To maintain a good global fit in the $L^1$ sense, we perform the mixture fit in linear space, rather than log-probability space: asymptotic errors in the tail distribution are simply not important.

We will introduce some notation to refer to our Gaussian mixtures. Let $N$ be the number of mixture components. Each component $k$, for $1 \leq k \leq N$, has a weight $w_k$, a standard deviation $\sigma_k$, and a mean $\mu_k$. Our approximation is therefore
\begin{align}
    \hat{\rho}(x) = \sum_{k=1}^N w_k\hat{\rho}^{(k)}(x)=\sum_{k=1}^N \frac{w_k}{Z_k}\exp\left(-\frac{(x-\mu_k)^2}{2\sigma_k^2}\right)
\end{align}
where we use the truncated Gaussian normalizer
\begin{align}
    Z_{k} = \int_0^\infty \exp\left(- \frac{(z-\mu_k)^2}{2\sigma_{k}^2}\right)\, dz = \sqrt{\frac{\pi}{2}}\sigma_{k} \operatorname{erfc}\left(-\frac{\mu_k}{\sigma_{k}\sqrt{2}}\right)
\end{align}

\subsubsection{Factoring in Thermal Error}
\label{sec:thermal_theory}

As we will see in \Cref{sec:exp_thermal}, thermal error can be effectively modeled as an independent zero-mean Gaussian perturbation of the output-spin pre-activations. Empirical modeling shows that the variance of the perturbations $\epsilon^2_\delta$ is governed by a power law, $C_T \delta^{\alpha_T}$. The Gaussian mixture approximation for the feed-forward pre-activations makes it very easy to estimate the distribution of spin averages under thermal error perturbation: it is simply $\tanh_* \hat{\rho}_\delta$, where $\hat{\rho}^{(k)}_\delta \sim \mathcal{N}(\mu_k, \sigma^2_k + C_T \delta^{\alpha_T})$. For computing error rates at finite $\delta$, we will use the blurred `pre-activation' distribution $\hat{\rho}_\delta$, which is also a Gaussian mixture. We will also use the notation $\sigma_{k,\delta}^2 = \sigma_k^2 + C_T\delta^{\alpha_T}$ for brevity. 

\subsection{Modeling the Stochastic Error Rate}
\label{sec:error_rate}

Let us first consider the probability of error on a single output spin $\sigma_i$.
After discarding burn-in, write $G$ for the number of retained sweeps and let
\begin{align}
    \langle \sigma_i\rangle_G
    :=
    \frac{1}{G}\sum_{t=1}^G \sigma_i^{(t)},
\end{align}
where $\sigma_i^{(t)}$ denotes the value of the spin $i$ at sweep $t$, and let $\mu_i$ denote the equilibrium mean of $\sigma_i$ under the Boltzmann distribution, i.e. $\langle\sigma_i\rangle_\infty$. 
An error occurs when $\operatorname{sign}\langle \sigma_i\rangle_G \neq \operatorname{sign}(\mu_i)$.
Since $\sigma_i\in\{-1,+1\}$, its equilibrium variance is $1-\mu_i^2$. We use the MCMC convention for the integrated autocorrelation time,
\begin{align}
    \tau_{i,\mathrm{int}}
    :=
    1 + 2\sum_{\ell=1}^{\infty}\frac{
        \operatorname{Cov}(\sigma_i^{(0)},\sigma_i^{(\ell)})
    }{
        \operatorname{Var}(\sigma_i)
    }.
\end{align}
With this convention, the Markov chain central limit theorem gives
\begin{align}
    \sqrt{G}\left(
        \langle \sigma_i\rangle_G - \mu_i
    \right)
    \xlongrightarrow{\mathcal{D}}
    \mathcal{N}\!\left(
        0,\,
        (1-\mu_i^2)\tau_{i,\mathrm{int}}
    \right),
\end{align}
or equivalently, for large $G$,
\begin{align}
\label{eq:singlespinerror_clt}
    \langle \sigma_i\rangle_G
    \approx
    \mathcal{N}\!\left(
        \mu_i,\,
        \frac{1}{G}(1-\mu_i^2)\tau_{i,\mathrm{int}}
    \right).
\end{align}
Therefore, by the Markov chain central limit theorem,
\begin{align}
\label{eq:singlespinerror}
    \mathbb{P}\!\left[
        \operatorname{sign}\langle\sigma_i\rangle_G
        \neq
        \operatorname{sign}(\mu_i)
    \right] \approx
    \frac{1}{2}
    \operatorname{erfc}\!\left(
        |\mu_i|
        \sqrt{
            \frac{G}
            {
                2(1-\mu_i^2)\tau_{i,\mathrm{int}}
            }
        }
    \right).
\end{align}

\subsubsection{Asymptotic Total Error Rate}
\label{sec:totalR}

\begin{figure}
    \centering
    \includegraphics[width=\linewidth]{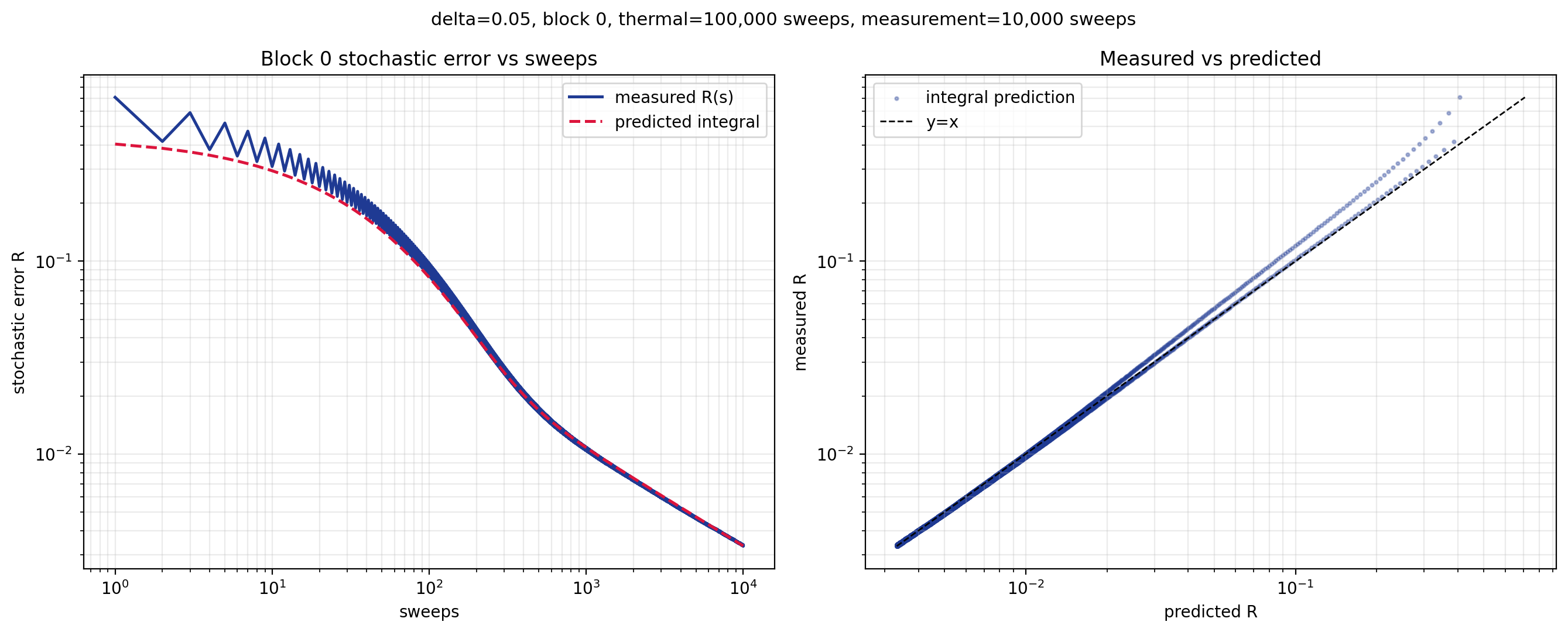}
    \caption{Measured stochastic error rate compared against numerically evaluated integral for the first block of our CIFAR-10 model. The fit is very close, but the integral does not capture the period-two oscillation caused by the chromatic Gibbs sampling.}
    \label{fig:stochastic}
\end{figure}

We have calculated the stochastic error rate for a single output spin as a function of its true mean and integrated autocorrelation time. Now, we extend our calculation of the stochastic error rate to the entire system, calculating $R$ by integrating our per-spin stochastic error rate over the distribution $\rho$. We will assume for simplicity that all spins have the same integrated autocorrelation time $\tau$. Then, we will start by writing the integral approximation,
\begin{align}
\label{eq:exact_integral}
    \mathbb{E}[R(G,\delta)] \approx \frac{1}{2}\int_0^\infty \text{erfc}\left(\sqrt{\frac{G}{2\tau}} \frac{\tanh(\delta z)}{\sqrt{1-\tanh(\delta z)^2}}\right)\hat{\rho}_\delta(z)\, dz = \frac{1}{2} \int_0^\infty \text{erfc}(\theta \sinh(\delta z)) \hat{\rho}_\delta(z)\, dz,
\end{align}
where we make the substitution
\begin{align}
    \theta := \sqrt{\frac{G}{2\tau}}.
\end{align}
An experimental validation, plotted in \Cref{fig:stochastic}, shows that this integral is an excellent approximation of the real stochastic error rate---though it cannot capture the period-two oscillation in error rates, which is an artifact of the two-color Gibbs sampling, a feature which is not detected by the CLT-style error estimate. At this point, we can easily find the asymptotic approximation of the tail for any $\delta$:
By substitution, we get 
\begin{align}
    \frac{1}{2\delta \theta} \int_0^\infty \text{erfc}(s) \hat{\rho}_\delta\left(\frac{1}{\delta}\text{arsinh}\left(\frac{s}{\theta}\right)\right)\frac{ds}{\sqrt{1+s^2/\theta^2}}
    = \frac{1}{\sqrt{\pi}}\int_0^\infty e^{-s^2} \hat{P}_\delta\left(\frac{1}{\delta}\text{arsinh}\left(\frac{s}{\theta}\right)\right)\, ds
\end{align}
where $\hat{P}_\delta$ is the CDF of $\hat{\rho}_\delta$. Letting $u = \delta\theta$, Taylor expansion then gives us the leading-order asymptotic formula
\begin{align}
\label{eq:asymp}
\mathbb{E}R = \frac{\hat{\rho}_\delta(0)}{2u\sqrt{\pi}} + \mathcal{O}(u^{-2}).
\end{align}
If $\hat{\rho}_\delta$ has an even differentiable extension through the origin, then $\hat{\rho}_\delta'(0)=0$ and the remainder improves to $\mathcal{O}(u^{-3})$. The primary asymptotic quantity is $u = \delta\theta$, which is proportional to $\delta \sqrt{G}$, obtaining the first-order power law approximation we promised in \Cref{sec:stochastic_error_definition}. Note that while $\hat{\rho}_\delta(0)$ does depend on $\delta$, it is only weakly dependent at small values of $\delta$, and this effect can be ignored for the purposes of understanding $\delta \rightarrow 0$ behavior. 

\subsubsection{Low-$\delta$ Closed Form}

We would like to understand how fast the rate $R$ converges to its leading power law. Introducing the magic approximation \cite{LETHER1993573}
\begin{align}
    \operatorname{erfc}(x) \approx e^{-q(x)} ,&& q(x) = \frac{16}{23}x^2+\frac{2}{\sqrt{\pi}}x,
\end{align}
and using the fact that for $\delta < \frac{1}{4}$ we can essentially ignore the $\sinh$ and obtain a nearly tight upper bound, we can approximate the integral from \Cref{eq:exact_integral} as
\begin{align}
    \int_0^\infty \text{erfc}(\theta \sinh(\delta z))\hat{\rho}_\delta(z)\, dz \leq \int_0^\infty \text{erfc}\left(uz\right)\hat{\rho}_\delta(z)\, dz\\
    \approx \sum_{k=1}^N \frac{w_k}{Z_{k,\delta}} \int_0^\infty \exp\left(- \frac{16}{23} u^2z^2 - \frac{2}{\sqrt{\pi}} uz- \frac{(z-\mu_k)^2}{2\sigma_{k,\delta}^2}\right)dz.
\end{align}
This is now a straightforward, if messy, application of the error function. By introducing the following shorthands,
\begin{align}
 a_k(u)
 :=
 \frac{64}{23}u^2+
 \frac{2}{\sigma_{k,\delta}^2},
&&
 b_k(u)
 :=
 \frac{2}{\sqrt{\pi}}u
 -
 \frac{\mu_k}{\sigma_{k,\delta}^2},
\end{align}
we can evaluate the integral for the Gaussian mixture component $k$ as
\begin{equation}
\sqrt{\frac{\pi}{a_k}}
\exp\left(
-\frac{\mu_k^2}{2\sigma_{k,\delta}^2}
\right)
\operatorname{erfcx}\left(
\frac{b_k}{\sqrt{a_k}}
\right).
\end{equation}
We then obtain a closed-form approximation for $\mathbb{E}R$,
\begin{align}
    \mathbb{E}R \approx \sum_{k=1}^N w_kR_k ,&& R_k =  \hat{\rho}_\delta^{(k)}(0)\sqrt{\frac{\pi}{16a_k}} \operatorname{erfcx}\left(
\frac{b_k}{\sqrt{a_k}}
\right).
\end{align}
The large-$u$ behavior of $R_k$ is governed by $a_k^{-1/2}$, which is asymptotically proportional to $u^{-1}$. The erfcx provides a small-$u$ correction factor, and becomes a $k$-independent constant at large $u$. It should be noted that the normalization ensures that $R_k(0) = \frac{1}{2}$.

\subsubsection{Convergence Rate}

\begin{figure}
    \centering
    \includegraphics[width=\linewidth]{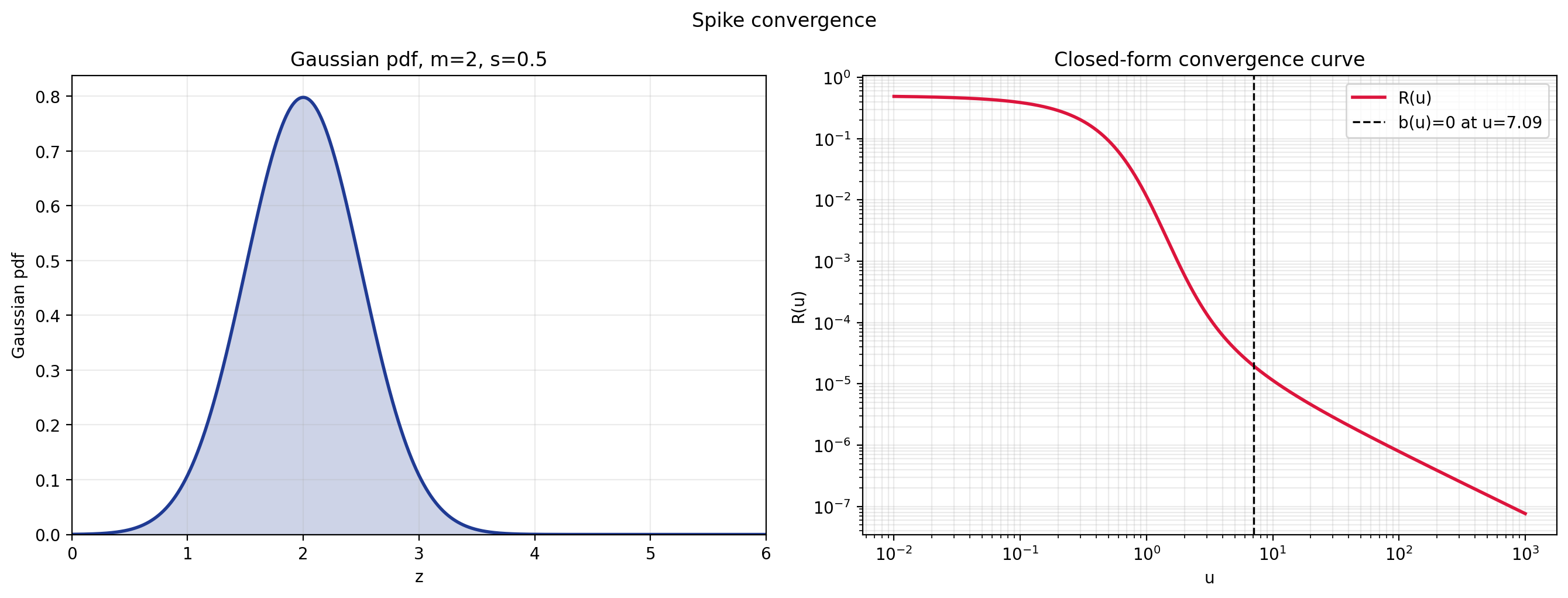}
    \caption{Theoretical convergence rate for a single Gaussian spike away from $z=0$, with time of transition to power-law convergence visualized.}
    \label{fig:spike_convergence}
\end{figure}

The closed-form solution to the integral allows us to inspect analytically how each spike in the distribution $\rho$ contributes to the convergence rate. The error rate contribution for a single Gaussian spike, as a function of $u$, follows a predictable structure: For very small $u$, the error rate decays from $1/2$ to $1/5$ nearly linearly. Then, as $u$ grows, the error rate decays roughly exponentially, as the bulk of the erfc kernel passes through the spike, and we see very fast convergence. At some point a crossover is reached where the erfc kernel is concentrated around the intersection of the spike with the $y$ axis, and past this point, the error rate stabilizes and decays like a $1/u$ power law. On a log-log plot, this results in a sharp fall-off in error which suddenly stabilizes into a straight-line descent. This behavior, along with the example spike in question, is depicted in the graph in \Cref{fig:spike_convergence}. In the case that the spike is centered on zero, there is no exponential-decay phase, and the error rate becomes a power law immediately. 

We would like to know how large $u$ must be before the $R$ curve converges to its long-term power law behavior.
Since we can approximately decompose $\rho$ into a sum of spikes, we can answer this question independently for each Gaussian spike. The key to understanding the cliff in the log-log plot is the contribution of the erfcx function: for large $u$ it converges to a constant, but for small $u$ it can contribute a massive pre-factor. This is because for $s>0$, $\operatorname{erfcx}(-s)\approx 2e^{s^2}$: thus, when $b(u) < 0$, the erfcx term is decaying nearly exponentially. It follows, then, that this decay stops when $b=0$, i.e.
\begin{align}
    u = \frac{\mu\sqrt{\pi}}{2\sigma^2}.
\end{align}
This implies that the contribution to the error rate of a Gaussian spike begins to closely approximate the power law when
\begin{align}
    G \geq \frac{\tau \mu^2\pi}{2\sigma^4 \delta^2}.
\end{align}
Beyond this point, the error rate contribution of the spike converges to the power law more slowly, with relative error order $u^{-1}$. Before this point, the error drops off much more quickly with increasing $G$. This crossover point can be quite large for a concentrated spike located away from the origin. From this, we can derive the overall qualitative behavior of $R$ in terms of the spike components of $\hat{\rho}$: for very small $u$, each spike contributes nearly equally. As $u$ grows, however, concentrated spikes far from the origin decay very quickly, while wide spikes near the origin decay much more slowly. In other words, the concentrated spike far from the origin quickly ceases to matter relative to the wide bulk near the origin. Eventually, all spikes contribute a power law, but the power-law contribution is proportional to the spike's mass at $z=0$; it follows that concentrated spikes far from the origin contribute almost nothing to the long-term power law rate. This means that the bulk of the error rate is contributed by the spikes with large intersection with the $y$-axis: the bulk spike centered at $z=0$ is predominant, as it has no exponential-decay phase and is a power law from the start, and any spike with $\sigma > \mu$ contributes significantly.

\section{Experiments}
\label{sec:experiments}

Now that we have developed theoretical tools for analyzing our thermodynamic blocks, we will apply them to analytical experiments demonstrating how to characterize and tune the trained blocks. We begin by showing the results of naive hyperparameter sweeping, exhibiting a heatmap of classification accuracy as a function of uniformly chosen $G$ and $\delta$, in \Cref{sec:exp_heatmap_uniformG}. This should give the reader a good qualitative picture of the capabilities of our thermodynamic models. Then, we move on to finer analysis. In \Cref{sec:exp_thermal}, we fit the Gaussian power law model described in \Cref{sec:thermal_theory} to experimentally measured thermal error. Then, in \Cref{sec:exp_stochastic_only}, we use our model of stochastic error to experimentally measure the power-law relationship between block-level stochastic error and overall classification accuracy. Then, we combine our analytical tools and modeled quantities in \Cref{sec:exp_adaptive}, demonstrating how we can predict model performance as a function of $\delta$ and $R$, and how a desired error rate can be converted to an implementable sweep-count schedule. Finally, we investigate the importance of the Gibbs regularization algorithm in shaping the performance heatmaps by performing an ablation experiment in \Cref{sec:exp_lambdafp}.

\subsection{Uniform-Sweep Inference Performance}
\label{sec:exp_heatmap_uniformG}

\begin{figure}[h]
    \centering
    \includegraphics[width=0.7\linewidth]{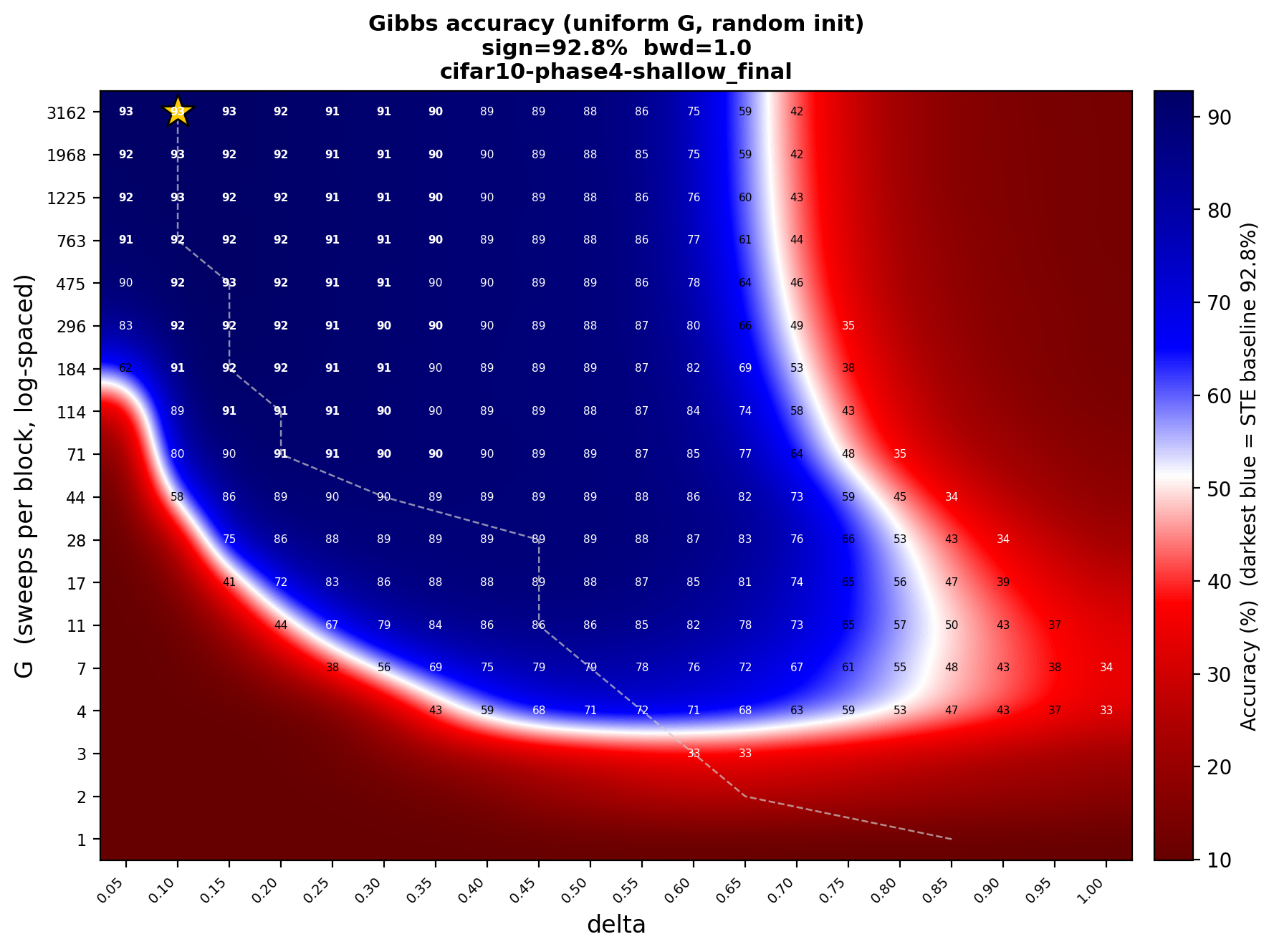}
    \caption{Gibbs accuracy on a $(\delta, G)$ grid with uniform $G$ per
    block, random initialization. The color map is centered on the STE
    baseline, so cells rendered in dark blue match STE accuracy and cells in
    dark red collapse to random. The dashed ridge marks the
    accuracy-optimal $\delta$ at each $G$; it drifts from large $\delta$
    at small $G$ (short-chain regime) toward small $\delta$ at large $G$
    (equilibrium regime).}
    \label{fig:uniformG_heatmap}
\end{figure}

Before we engage in more sophisticated methods for selecting $G$ given $\delta$, it is worth seeing how the two hyperparameters determine classification accuracy directly. The simplest possible inference schedule is to set $G$ uniformly for all blocks. To visualize the qualitative effects of different choices of $G$ and $\delta$, we plot accuracy against these two variables in \Cref{fig:uniformG_heatmap}, for the case of our 6-block CIFAR-10 model. This plot visually summarizes several of the key features that we have been discussing: increasing cost-to-converge as $\delta \rightarrow 0$, convergence to correctness for $\delta < \delta_\text{train}$, increasing performance ceiling as $\delta \rightarrow 0$, and poor performance for $\delta > \delta_\text{train}$ or $G$ not large enough. We would like to observe that mid-range $\delta$ can give surprisingly good performance at very low sweep count. An accuracy of $89\%$ can be achieved with $\delta=0.45$ using only 102 total Gibbs sweeps. This shows that if inference-cost minimization is the highest priority, we can still get pretty good results at very low sweep cost; the corresponding power consumption will depend on the hardware implementation.

\subsubsection{Explaining the High-$\delta$ Overfit}
\label{sec:overfit}

One feature of \Cref{fig:uniformG_heatmap} which immediately draws attention and hasn't been addressed by the theory discussed so far is the right-hand side of the diagram: for $\delta > \delta_\text{train}$, as the Markov chain progresses we see first an increase in accuracy, then a decrease back to random. It is probably not a coincidence that the peak accuracy in this regime occurs roughly at $G = G_\text{train}$. In other words, the Gibbs chains here have `overfit', i.e. they have learned only at the $G$ they were trained for, rather than generalizing to the equilibrium state. This is a satisfactory moral explanation, but it still does not mechanically explain this phenomenon, which is rather odd: how can a system reliably go from random, to close-to-correct, and then move to its real equilibrium, which is totally wrong? Somehow, both the random initial state and the final equilibrium are bad, but the system reliably passes through a `good' region on the way from one to the other. 

At $\delta = 0.75$, when we sweep long enough ($G>1000$, say) that the system is essentially converged, the accuracy is terrible. It follows that this poor accuracy is caused by excessive thermal error. Since the system produces better accuracy at 10 sweeps per block, it follows that the better accuracy is \textit{caused} by stochastic error. In other words, at high $\delta$, stochastic error can partially cancel out the effects of thermal error. However, as we will see in \Cref{sec:exp_adaptive}, this cannot be due to broad statistical effects, as injecting i.i.d. thermal and stochastic error cannot replicate the effect. In other words, this anomaly may be caused by spin-spin correlation patterns, something which our theory cannot capture. The fact that the effect occurs at high $\delta$, and therefore higher coupling strengths, strongly supports this inference. However, it should also be noted that this anomaly occurs away from the real region of interest: to get good performance, we want low $\delta$, and that is precisely where our independent-spin theory functions as expected. 

\subsection{Modeling Thermal Error}
\label{sec:exp_thermal}

\begin{figure}[h]
    \centering
    \includegraphics[width=0.6\linewidth]{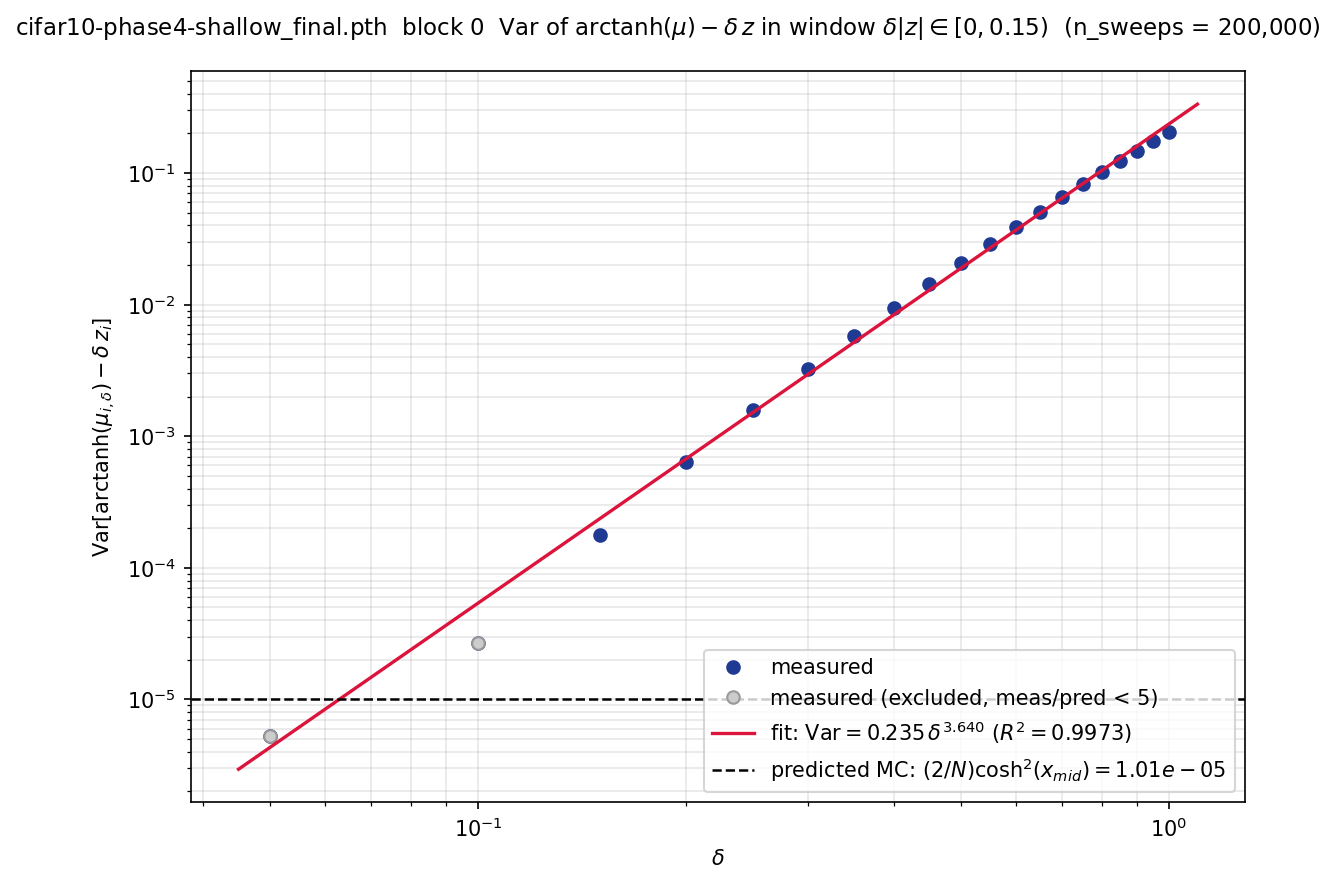}
    \caption{Experimentally measured variance of per-spin thermal perturbation, as a function of $\delta$; a power law fits the data cleanly. The dashed line indicates the estimated stochastic noise floor.}
    \label{fig:thermal_power_law}
\end{figure}

It is difficult to isolate thermal error because we cannot access the true spin averages $\mu_i$. However, we can just estimate $\mu$ by using a very large number of Gibbs sweeps, so that the stochastic error is very small, and assess the differences between $\mu_i$ and $\tanh(\delta z_2)$, with the understanding that some noise has entered into the measurement. While this relationship is quite complex and difficult to capture theoretically, it turns out that we have a pretty good approximation using independent Gaussian noise,
\begin{align}
\label{eq:thermal_perturbation}
    \arctanh(\mu_{i,\delta}) - \delta z_i \approx \mathcal{N}(0,\epsilon_\delta^2).
\end{align}
In other words, we can essentially model thermal error as a Gaussian perturbation of pre-activations in the feed-forward model. This approximation is not perfect. In practice $\epsilon$ isn't actually uniform in $z$, and the distribution isn't actually Gaussian (it has some non-negligible excess kurtosis), but this is a good enough approximate model. We mostly care about thermal error when it affects spins whose pre-activation is close to zero: these spins are the most likely to introduce sign errors due to thermal perturbation, while spins with strong pre-activation are not likely to change sign under thermal perturbation.

In order to use this model, we need to actually measure $\epsilon^2_\delta$. We experimentally measure the pre-activation perturbation defined in \Cref{eq:thermal_perturbation} for various values of $\delta$, specifically at spins where $|\delta z_i|<\frac{1}{4}$. We then obtain the variance of the perturbation, and fit a power law. The result, shown in \Cref{fig:thermal_power_law}, shows that a power-law fit is very close over the measured range. Using this power law, we can now inject simulated thermal error at any $\delta$ by perturbing pre-activations according to random Gaussian noise. This will be our model for \textit{synthetic thermal error} and is used in the following experiments. 

\subsection{Global Effects of Stochastic Error}
\label{sec:exp_stochastic_only}

\begin{figure}[h]
    \centering
    \includegraphics[width=\linewidth]{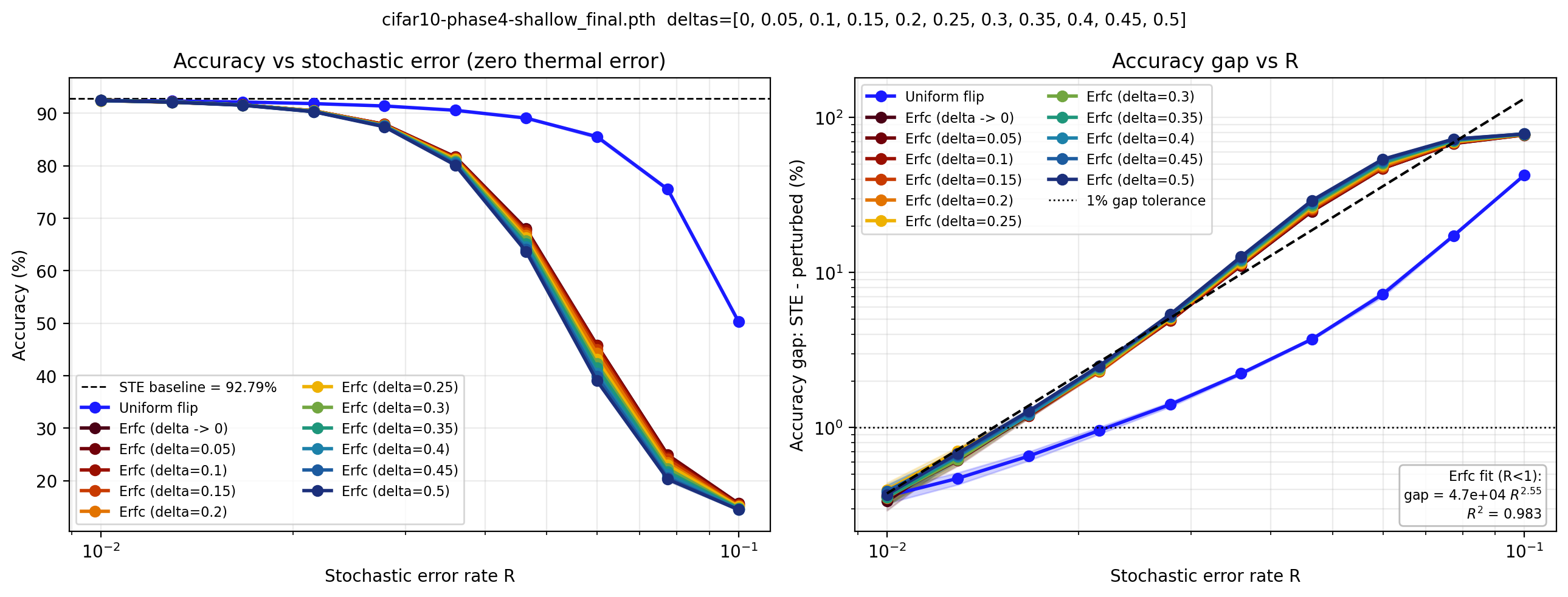}
    \caption{Accuracy versus synthetic stochastic error at zero thermal error fits well to a power law.}
    \label{fig:stochastic_only}
\end{figure}

It will be useful to see what the direct relationship between stochastic error and overall accuracy is, assuming thermal error is zero and ignoring the specifics of the Gibbs dynamics. We will synthetically inject stochastic error $r$ at each block and run inference in feed-forward mode, then measure accuracy. Naively, we could just flip spins i.i.d. uniformly, but our analysis shows that stochastic error is not evenly distributed: it affects small-mean spins more frequently, and the weighting depends on $\delta$. Suppose we have selected $\delta$ and the target error rate $r$. Then, using binary search, we find $\gamma$ such that 
\begin{align}
\label{eq:stochastic_synthetic}
 r=\frac{1}{|s_2|}\sum_{i=1}^{|s_2|} \frac{1}{2}\text{erfc}\left(\gamma \sinh(\delta |z_i|)\right),
\end{align}
where $z_i$ is the feed-forward pre-activation of output spin $i$. In the case that $\delta = 0$, we use the $\delta \rightarrow 0$ asymptotic limit by instead performing binary search on 
\begin{align}
    r=\frac{1}{|s_2|}\sum_{i=1}^{|s_2|} \frac{1}{2}\text{erfc}\left(\gamma|z_i|\right).
\end{align}
In either case, once the right $\gamma(r)$ is determined, the terms inside the sum are the probabilities that each spin should be flipped. We can easily sample these flips independently. 

We perform an experiment on our CIFAR-10 model by injecting stochastic error at rate $R$ into the feed-forward model at zero thermal error. The results are depicted in \Cref{fig:stochastic_only}. Uniform flipping is actually very optimistic: in fact, correctly weighted injected stochastic error degrades performance much more quickly than uniform error. This leads us to the understanding that low-$|\mu|$ output spins are disproportionately important for classification performance. This explains the results in other experiments which show that accuracy is essentially worthless as $G$ increases, up until a sharp phase transition to near-perfect accuracy, which we infer must occur when the low-$|\mu|$ spins begin to converge. Interestingly, we see clear power-law behavior relating the stochastic error rate to the accuracy gap at low values of $R$. A linear regression shows that the accuracy loss is roughly proportional to $R^{2.55}$.

\subsection{Optimal Selection of Sweep Count}
\label{sec:exp_adaptive}

\begin{figure}[h]
    \centering
    \includegraphics[width=0.7\linewidth]{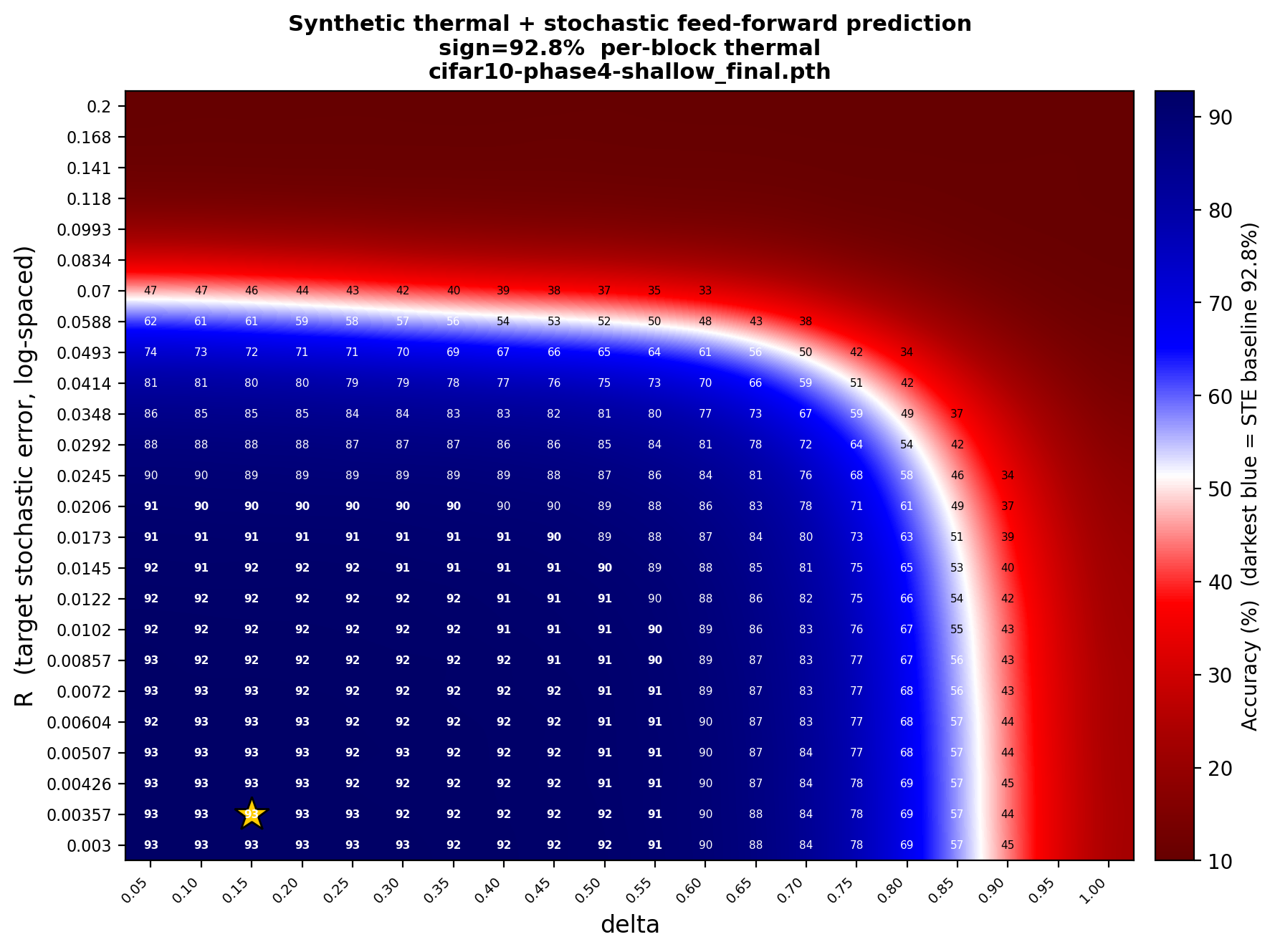}
    \caption{Theoretical performance of CIFAR-10 classifier at varying $R$ and $\delta$, based on synthetic stochastic and thermal noise.}
    \label{fig:theory_performance}
\end{figure}

How do we select the number of sweeps to run our models with? Naively, we might use the same number of sweeps for each block, and simply increase the sweep count until we get good performance. This is the approach depicted in \Cref{fig:uniformG_heatmap}. However, this is not efficient, nor is it sensitive to the problem: some blocks are more expensive to run than others, and ideally we would allocate more sweeps where they are needed, thus minimizing cost while maximizing performance. Instead of specifying $G$ upfront and seeing what performance comes out, we would like to specify performance up front and see what $G$ comes out. 

We can do exactly that using our models of stochastic and thermal error. To begin with, we can use our idealized synthetic noise models to theoretically predict network performance at varying levels of $R$ and $\delta$. To do this, we run the model in feed-forward mode, injecting noise while we do so. First, we inject synthetic thermal noise based on $\delta$, perturbing the pre-activations of the output layers based on \Cref{eq:thermal_perturbation}. Then, we flip spins at a rate $R$ using the relationship \Cref{eq:stochastic_synthetic}. We plot the resulting classification accuracy from this experiment in \Cref{fig:theory_performance}. This prediction captures the overall performance characteristics of the network, though it is not sensitive to more complex and subtle behavior like the large-$\delta$ overfit: note that at large $\delta$, it fails to predict the bad-good-bad transition with decreasing $R$. This is likely due to the fact that the overfitting behavior is caused by underlying spin-spin correlations and delicate structure of the energy function which cannot be captured by the independent-spin synthetic noise model. 

\begin{figure}[h]
    \centering
    \includegraphics[width=\linewidth]{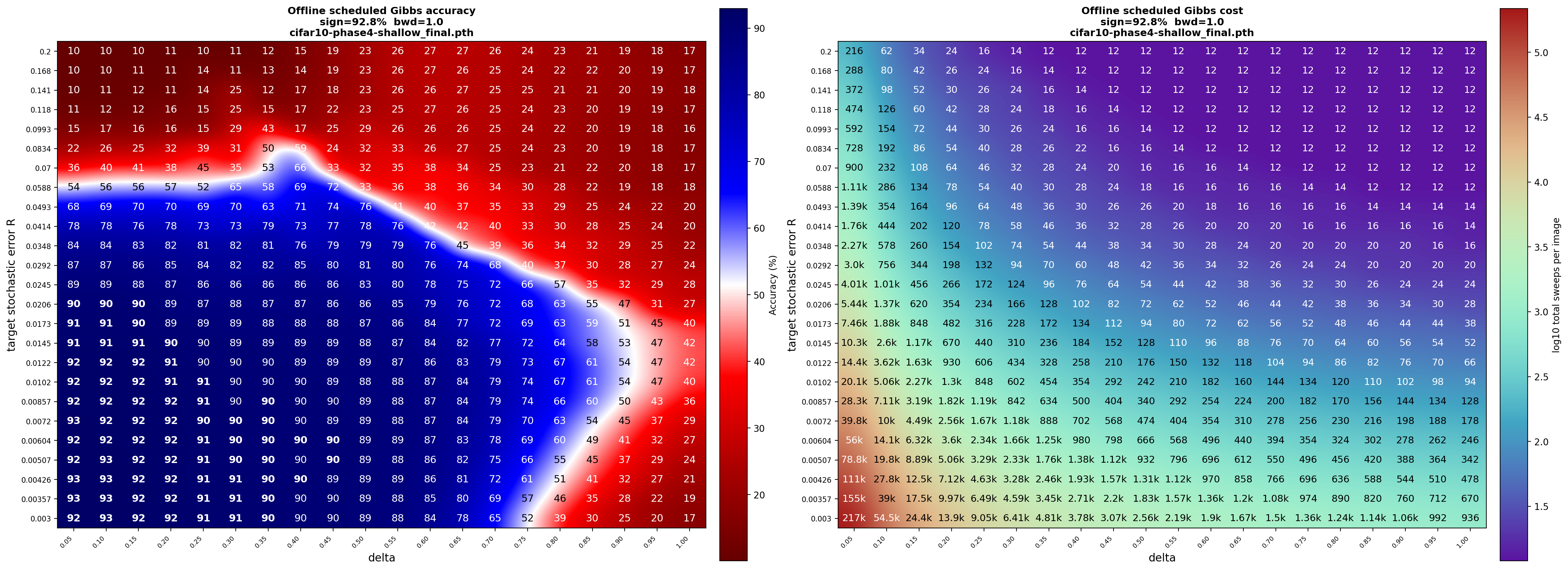}
    \caption{Experimental performance for varying $R$ and $\delta$, with sweep counts selected by our theory. The left plot shows classification accuracy; the right plot shows total sweep cost.}
    \label{fig:offline}
\end{figure}

Now, we will test these predicted accuracies with an experiment. In order to do this, we need to convert a target value of $R$ to a prescription for sweep count on each block. First, we fit a Gaussian mixture model to the empirical histogram of pre-activations, as described in \Cref{sec:gaussian}. Then, we compute $G(R)$ by performing binary search on the numerical integral \Cref{eq:exact_integral}, with the additional correction for the average integrated autocorrelation time $\tau = 1 + 0.71\delta^2$, based on our eigendecomposition theory (see \Cref{sec:autocorrelation}). This converts a pair $(\delta, R)$ to a sweep-count schedule for each block. The results of performing the experiment on our CIFAR-10 model are plotted in \Cref{fig:offline}. We show both the accuracy heatmap and a total sweep cost heatmap. As expected, we see that lower $\delta$ and lower $R$ result in better performance and higher cost. 

While the experimental results in \Cref{fig:offline} overall confirm the theoretical prediction depicted in \Cref{fig:theory_performance}, there are some discrepancies to take note of. First of all, we note that the experimentally measured good-bad boundary for $\delta < 0.7$ is rougher and less uniform than the theoretical prediction. For instance, there is an unexpected bump in accuracy around $\delta =0.4, R=0.07$. We note, however, that the total sweep cost for this measurement is only 24, meaning an average of 4 sweeps per block. By comparison with \Cref{fig:uniformG_heatmap}, we see that for small sweep counts, one more or less sweep can have a very significant effect on accuracy, especially at mid-range $\delta$. Therefore, the relationship of estimated error to realized error becomes quantized at low costs, resulting in a significantly more jagged pattern: we may desire a certain stochastic error level, but we are forced to only do an integer number of sweeps. 

The second discrepancy between the two plots is more fundamental: the shape of the blue region beyond $\delta = \delta_\text{train} = 0.7$ does not curve down smoothly as in the theoretical plot, but instead bumps outwards before curving back. This is precisely the high-$\delta$ `overfitting' effect that we discussed in \Cref{sec:overfit}. The fact that it appears in the experiment, but not in the theory, is very telling regarding the etiology of the overfitting effect: the theoretical model does not consider spin-spin correlation patterns, instead treating the statistical aggregate as composed of independent errors. This assumption causes the overfitting effect to disappear. We take this as evidence consistent with the overfitting effect being caused by spin-spin correlation patterns and the non-random interaction of thermal and stochastic error effects. This makes sense as a hypothesis from a theoretical point of view, since the high-$\delta$ regime is precisely that mode of operation in which the spin-spin couplings are strongest, and the independent-spin mean field assumption that we used to construct the theory is least accurate. However, on a practical note, we note that the best accuracy is achieved at low $\delta$, which is precisely where our theory is most effective. In other words, our theory works where we need it to work.

\subsection{Regularizer Strength: $\lambda_{\mathrm{FP}}$ Tradeoff}
\label{sec:exp_lambdafp}

\begin{figure}[h]
    \centering
    \includegraphics[width=0.7\linewidth]{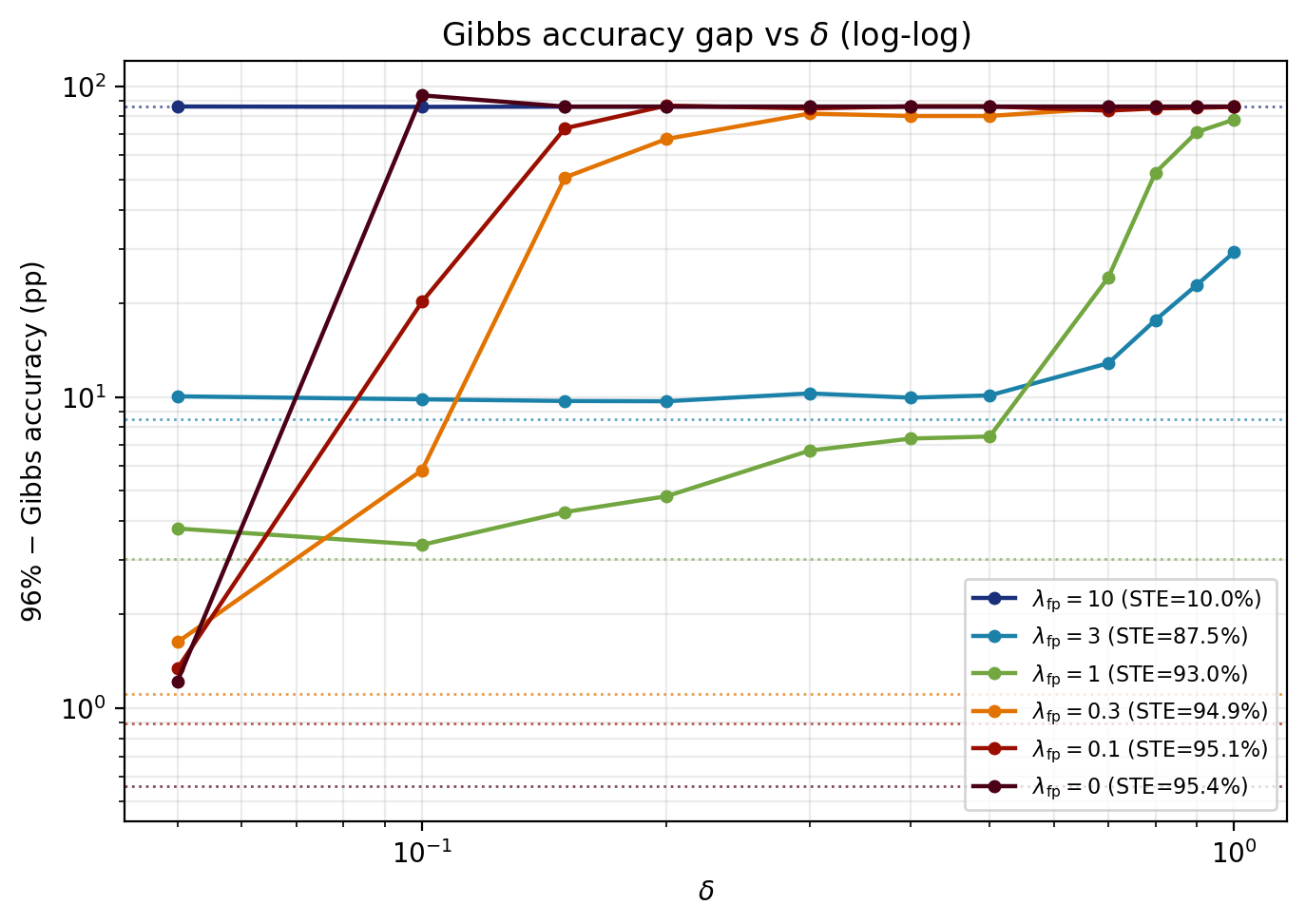}
    \caption{Log-scaled plot showing error rates against $\delta$ for a network trained with varying values of $\lambda_\text{FP}$. As expected, lower values of $\lambda_\text{FP}$ lead to better performance at small $\delta$, but faster accuracy collapse as $\delta$ increases. For $\lambda_\text{FP}$ too high, training collapses completely.}
    \label{fig:lambdafp_tradeoff}
\end{figure}

The parameter $\lambda_\text{FP}$ controls how strong the Gibbs regularization is. Since this regularization term is the most critical part of the training algorithm, the selection of $\lambda_\text{FP}$ is vital to achieving the desired results. The key tradeoff is between STE accuracy and Gibbs robustness: small $\lambda_\text{FP}$ yields a higher accuracy ceiling, but demands higher $G$ and lower $\delta$ to approximate it. Large $\lambda_\text{FP}$, on the other hand, produces a robust network which can operate at higher $\delta$ and fewer sweeps, but has a lower maximum performance cap. 

We can view this tradeoff quite cleanly by doing an ablation experiment: we will train our 6-block CIFAR-10 model for several different values of $\lambda_\text{FP}$, and plot its performance and cost when inference is performed with adaptive sweeping at a fixed value of $R$. The results are plotted in \Cref{fig:lambdafp_tradeoff}. The experiment used $R = 0.01$ and $\delta_\text{train} = 0.7$. We observe the expected behavior: lower $\lambda_\text{FP}$ leads to better maximum performance at small $\delta$, but also a quicker falloff in accuracy as $\delta$ increases. Larger $\lambda_\text{FP}$ leads to flatter performance, i.e. better resilience to larger $\delta$, but also lower overall performance. Accuracy always falls off beyond $\delta = \delta_\text{train}$ in any case, as expected. The extreme values of $\lambda_\text{FP}$ are also worth noting: $\lambda_\text{FP} = 0$ (regularizer disabled) does produce great accuracy at very small $\delta$, leaning entirely on the magnitude regularizer, but this performance evaporates entirely with even slightly larger $\delta$. Extremely large $\lambda_\text{FP}$, on the other hand, predictably causes total training collapse. 

% ======================================================================
\section{How Close Are We to Hardware?}
\label{sec:hardware}

This work is a hardware-aware software paper. Its purpose is to develop algorithms for an emerging class of probabilistic and thermodynamic devices, not to deploy them on currently available hardware. Our experiments address a key gap in algorithmic research: can we scale the training of models capable of running on Ising hardware to practical sizes? Our results show that pure backpropagation training is capable of producing large-scale deep models which function under Gibbs sampling and binary inter-block communication, fulfilling key requirements for the development of software for Ising-type systems. If Ising hardware systems continue to become bigger and more powerful, our results indicate that the model-training algorithms can scale alongside that hardware advancement. 

However, stability under Gibbs sampling and binary I/O are not the only requirements common to all existing Ising hardware. In practice, almost every Ising hardware system also imposes some restrictions on weight quantization and sparsification, often requiring specific graph topologies. Currently, a wide variety of experimental hardware options exist. Some technologies promise dense graph topology, but extreme weight quantization, while some promise high weight precision at the cost of very restrictive graph topologies. The approaches differ greatly in technology readiness level (TRL), roadmap, and forecasted advantages. In short, a single ``Ising hardware'' standard simply does not exist: what does exist is a diverse family of prototypes, each with its unique caveats and tradeoffs. We will first briefly review the landscape of currently available options, then discuss broadly how quantization and sparsification can be added to our models. 

\subsection{Review of Current Hardware}

At present, there is no single standard Ising-machine architecture. The phrase ``Ising machine'' covers several families of systems whose capabilities are not directly comparable. Some are physical annealers, some are optical or spintronic stochastic systems, and others are quantum-inspired digital or FPGA/GPU solvers. Some operate on stochastic silicon, while some require more exotic materials or environments. This diversity is encouraging, but it also means that ``running on an Ising machine'' is not a well-defined hardware target. The details of many machines are trade secrets, and the landscape is not always clear. However, we will review some of the existing publicly-known systems to understand the general sense of scale and the available tradeoffs. 

\subsubsection{Superconducting Quantum Annealers}
D-Wave systems are the most mature commercial examples of physical Ising/QUBO-oriented hardware. The current Advantage2 system has thousands of qubits, tens of thousands of couplings, and a sparse Zephyr topology with bounded local connectivity \cite{dwave_advantage2_docs,dwave_advantage2_release}. These machines are highly relevant to the broader Ising-computing ecosystem, but they are not the most relevant target for the current paper. Quantum annealers are expensive and energy intensive, leaving little economic justification for running AI inference on them instead of a classical GPU. Besides that constraint, they are also designed, as the name implies, for \textit{annealing} to low energy states---a mode of operation incompatible with the high-temperature sampling discussed in this paper. On the other hand, a direct connection between quantum computing and AI models may indeed yet be useful, and the sampling-temperature issue may not be unresolvable.

\subsubsection{Digital and CMOS Ising Systems}
Digital annealing systems, including Fujitsu's Digital Annealer and Hitachi's CMOS/Momentum Annealing line, provide a different tradeoff. They use conventional semiconductor hardware or hybrid software-hardware systems to solve large QUBO or Ising optimization problems with high parallelism \cite{fujitsu_da_benchmark_2025,hitachi_relaxed_ma_2024}. Digital platforms can handle much larger logical problem sizes than sparse physical quantum annealers, and some support dense or effectively dense QUBO formulations. This is attractive from the perspective of variable count and connectivity. These platforms, as the name suggests, typically target annealing problems for combinatorial optimization, a typical traditional application of the Ising model for computation. As digital systems, the underlying technology is more malleable, and likely could be modified to support our models' sampling requirements. However, the exact technical details of such systems are not fully known: these systems provide a useful gauge of scale for existing hardware, but specific discussions of compatibility are speculative.

\subsubsection{Simulated Bifurcation Machines}
Toshiba's Simulated Bifurcation Machine and SQBM+ platform represent another important quantum-inspired direction. These systems implement Ising-like optimization through highly parallelizable simulated bifurcation algorithms on GPUs, FPGAs, and cloud infrastructure, and current public materials describe support for 10 million spins, and up to 2,000 spins with fully connected interactions \cite{toshiba_sqmplus_2026,toshiba_sbm_edgechaos_2026}. This makes the platform significant for large-scale Ising optimization. From the perspective of thermodynamic neural inference, the same caveat applies: simulated bifurcation is a solver for optimization dynamics, not a drop-in replacement for the fixed-temperature Gibbs time averages used in our blocks. Nonetheless, these machines are large-scale implementations of the Ising system, and are implemented on malleable substrates like FPGAs. In other words, this technology likely can support our models, but needs modifications to its harness to do so. 

\subsubsection{Coherent Ising Machines}
Coherent Ising machines use optical degrees of freedom, such as degenerate optical parametric oscillator pulses, to represent spins. Large measurement feedback systems have demonstrated very large spin counts and dense effective couplings, including $100{,}512$ spins and more than $10$B pairwise interactions in an NTT coherent Ising machine \cite{ntt_cim100k}. This scale is impressive and relevant to the long-term hardware story; in fact, dense connectivity at these scales is far more than we need. However, current research focuses on sampling a single amplified spin configuration---in other words, Boltzmann time averaging would require a significant modification. Additionally, weight quantization requirements are extreme; coupling strengths on published machines are required to be in $\{-1, 0, 1\}$. Engineering a model to fit into this ternary quantization is not impossible, but would require tradeoffs and aggressive quantization-aware training algorithms. 

\subsubsection{Probabilistic-Bit and Spintronic Ising Machines}
The closest conceptual match to our model is the probabilistic-bit, or p-bit, family of machines. These systems implement tunable stochastic binary units and update them using coupling logic, making them naturally aligned with Gibbs-like sampling and Boltzmann-machine-style computation. Recent magnetic tunnel junction demonstrations show device-level progress, but current experimental systems remain much smaller than the networks studied here \cite{yang2026mtj_pim}. For this reason, p-bit and related spintronic probabilistic architectures may be the most natural long-term substrate for thermodynamic neural inference, even though they are not yet at the required scale.

\subsubsection{FPGA and GPU Gibbs Samplers}
Finally, classical FPGA and GPU implementations of Gibbs sampling are important baselines. They are not thermodynamic physical devices in the same sense as spintronic or optical systems, but they can implement the relevant Markov chain directly and can therefore directly test the algorithmic interface assumed in this paper. Prior Gibbs-sampled Ising classifiers such as \cite{Niazi_2024} are especially useful in this respect. They show that sweep-based Ising inference can be accelerated substantially by dedicated hardware, bringing a large number of sweeps within a reasonable cost range both in terms of energy expenditure and inference time. We discuss this comparison in more detail in \Cref{sec:sweepcount}.

\subsection{Sparsification and Quantization}
\label{sec:sparsification}

Practical Ising hardware and finite-precision Ising emulators typically require quantized couplings. Many also require strong sparsity or structured topology. As we have discussed, there is no standard for exactly how much quantization or sparsification is required. Therefore, we have no exact target to shoot for. However, since both ingredients are a key part of making thermodynamic AI systems practical on Ising machines, we will discuss in some generality how sparsification and quantization can factor into our algorithms. 

Although convolutional layers are sparse compared with fully connected layers, large channel counts still produce high local degree after unrolling. In the last block of our CIFAR-10 model, for example, the hidden layer has $784$ channels. A hidden spin connected by $3\times 3$ filters to both the input and output layers has local degree
\[
    2\cdot 3^2\cdot 784 = 14{,}112,
\]
which is the largest local degree in the model. Reducing this to maximum degree $141$ would require roughly $99\%$ sparsity, while reducing it to maximum degree $14$ would require roughly $99.9\%$ sparsity.

These are aggressive targets, beyond what should be expected from a purely post-training pruning pass. Such low spin degrees would entail integrating training techniques such as quantization-aware training, sparsity-aware training, structured channel sparsity, grouped convolutions, low-rank factorizations, and lottery-ticket-style dynamic sparsification \cite{evci2021rigginglotterymakingtickets}. These options are all natural extensions of the present pipeline. In the current paper, however, we intentionally keep the main algorithm simple and use a post-training compression experiment to demonstrate that our trained thermodynamic models are, in fact, amenable to quantization and sparsification, and not excessively brittle. 

We apply Wanda \cite{sun2024simpleeffectivepruningapproach} to the moderate-$\lambda_{\mathrm{FP}}$ six-block CIFAR-10 model. Before compression, this model has feed-forward STE accuracy $92.75\%$ and Gibbs inference accuracy $92.13\%$ at $\delta=0.1$. With $7$-bit weight quantization and $80\%$ sparsity, the feed-forward STE accuracy falls to $88.59\%$, while Gibbs inference accuracy at $\delta=0.1$ falls to $86.7\%$. The maximum per-spin degree is reduced to approximately $2.8$k. This suggests that while simple post-training approaches have their limitations, our models have some resilience to compression. Our Gibbs-stability tuning methods appear reasonably robust and do not decay much faster than overall feed-forward accuracy does under compressive pressure. While this result is not yet aggressive enough to fit our CIFAR-10 model onto current hardware, it is a good sign that more advanced methods are worth implementing, and that there is a clear roadmap for the training of compressed hardware-compatible models using our methods. 

\subsection{Sweep Count}
\label{sec:sweepcount}

Inference cost needs to be conceptually separated from hardware capacity. Capacity asks whether a device can represent the required spins, couplings, precision, and I/O pattern. Inference cost, however, refers to how much must be expended to obtain a result from running that device. In our model, inference cost is measured in terms of Gibbs sweeps, though hardware implementing other dynamics would have some other analogous measure of system simulation cost. Our convergence theory was motivated by the need to control inference cost. A model which works in theory but requires impractically high sweep count is not very useful. Therefore, we have quantified our sweep cost and shown theoretically that it does not diverge unexpectedly. However, the reader may still wonder how realistic the sweep counts listed in \Cref{tab:results} actually are in the current hardware context. 

For comparison, \cite{Niazi_2024} uses $10^5$ sweeps for MNIST and FashionMNIST-scale Gibbs-sampled Ising classification on an FPGA device, claiming that this number of sweeps can be executed within a few milliseconds and at low power consumption. Our smallest models use three orders of magnitude fewer sweeps, while our larger models still use substantially fewer sweeps than the previous result. This does not mean that any current device can run our large models within a few milliseconds, but it does show that sweep counts of order $10^5$ can be practical on at least one specialized implementation. Whether that timing transfers to our much larger models remains a hardware question, and power expenditure should be expected to depend on model size and architecture. In short, this comparison shows that sweep count does not by itself rule out practical implementation, and may be a point of improvement over existing results. However, without a fair and direct comparison, we cannot make any concrete and precise conclusions about the time or energy cost of running our models in real hardware. 

\subsection{Implications for Hardware Design}

The goal of thermodynamic AI research is to create useful models that can run on emerging probabilistic hardware. We have developed algorithms that can scale alongside Ising hardware as its capabilities expand and provided evidence that training algorithms need not be the immediate limiting factor for implementing AI models on thermodynamic systems. However, our algorithmic work is also important because it provides a target for hardware to shoot for. Thermodynamic AI will not become a reality at scale unless hardware research considers it a possibility and works towards moving Ising hardware in a direction that can support it. A common theme in our review of existing hardware is that most approaches have targeted the traditional application of Ising systems, annealing combinatorial optimization problems. This can lead to the development of more powerful Ising systems, but thermodynamic AI will require different adaptations of the underlying system to become a reality. To this end, we will elaborate the implications of our algorithmic research for the development of Ising hardware. 

The first implication is that future thermodynamic inference hardware should be developed to support convolutional graph topology. Convolutional inference, being naturally sparse, is a natural direction for Ising hardware. It also presents many promising applications for low-power edge inference, from implantable medical sensors to miniature surveillance systems, due to its natural applications in image, sensor, and audio processing. However, most existing hardware was not designed with this workload in mind. Hardware approaches which require specific graph topologies could be designed using an unrolled-convolutional network topology. On the other hand, digital systems could exploit weight sharing and translational symmetry to compress couplings in convolutional Ising layers in the same way that GPUs do. Since convolutional networks are a natural application of the technology, and training algorithms for programming convolutional Ising hardware now exist, it makes sense for convolutional weights to become a hardware-level priority. 

The second implication is that average-and-sign readout should be implemented as a hardware-native operation. Much Ising-machine work focuses on either ground-state readout or instantaneous Boltzmann samples. Our computation uses a different observable: the sign of the time average of an output spin. This can be computed in host software by collecting repeated Boltzmann samples, but doing so wastes bandwidth and moves a simple reduction away from the device. A thermodynamic neural chip should attach lightweight accumulation circuits to designated output spins, track their running averages over the sweep window, and export only the final signs or low-precision average estimates.

The third implication is that input spin fixing should be a first-class hardware feature. Throughout the paper, each thermodynamic block receives a binary input tensor and treats the corresponding input spins as fixed while the hidden and output spins evolve. Existing platforms can sometimes approximate this by applying large local fields, but this is not the same as a clean hardware-level clamping mechanism. Our neural inference naturally uses Boltzmann distributions conditional on the input spins, so a device intended for this workload should support input pinning and conditional Ising simulation. On many platforms, especially digital implementations, such a change could be straightforward, but it has to be kept in mind during the design process. 

% ======================================================================
\section{Conclusion}

In this paper, we have exhibited a backpropagation-based training algorithm for creating large convolutional networks which perform under block-wise Ising-system Gibbs sampling. The ability to train such a network at scale is a necessary building block for implementing AI models on Ising-type thermodynamic hardware. We have furthermore shown experiments and calculations demonstrating the relatively fast convergence and high performance of these models, as well as their ability to handle sizable benchmarks like CIFAR-100 with millions of spins. While not yet at the level of being uploaded directly to hardware, we have de-risked the production of Ising-compatible models and shifted the natural next focus towards enforcing device-specific hardware compatibility constraints. 

In other words, deep thermodynamic neural inference is now algorithmically viable enough to become a hardware-software codesign target. With the right training procedure and hardware implementation, Ising-compatible stochastic dynamics can support modern convolutional inference, and the cost of that inference can be analyzed, predicted, and designed rather than guessed. This paper takes the step from a theoretical correspondence between neural inference and stochastic systems toward a practical framework for scaling thermodynamic AI. That said, in this paper we have focused on arguing the core question of viability more than elaborating on every possible direction or pushing for detailed hardware implementation. Therefore, there are many natural directions open for future research that aims to flesh out, exploit, and realize the above results. 

Implementing a thermodynamic computer in real hardware requires two things. The software must be designed to fit the hardware specifications, and the hardware must support the demands of the software. To see the models we have discussed running on real Ising hardware, both sides must be addressed. First, a specific hardware platform must be picked as a codesign target. Then, the training algorithms must be refined to target that platform and its particular specifications. In reality, this would likely look like a modification to the existing procedure which integrates the right flavor of quantization, sparsification, and graph topology restriction. In short, specialization and algorithmic refinement are needed for implementing our models on any particular hardware platform.

Not all thermodynamic computing hardware is based on the Ising model. Some systems operate according to completely different principles, but another large class consists of hardware devices that implement a physical model closely related to but not identical to the Ising model. A common example is the Kuramoto model for coupled oscillators, implemented for example by coupled LC oscillator arrays with sub-harmonic injection locking \cite{11186164, shil}. In the low-temperature (i.e. strong locking) limit, the Kuramoto Hamiltonian has the same ground states as the analogous Ising Hamiltonian. However, more weakly coupled Kuramoto systems, possibly with injected stochastic perturbations or possibly with modified oscillators \cite{khan2025analyzingparametricoscillatorising}, approximate the Ising model inexactly. Nonetheless, the existence of prototype Kuramoto hardware indicates that it may be worth extending our mean-field theoretical correspondence. If we can replicate our methods for training thermodynamic hardware with cousins of the Ising model, the number of possible hardware application targets will increase. 

In the paper, we have focused only on using our convolutional networks for image classification. This was mostly for simplicity and uniformity. In fact, our methods should extend to a broad class of convolutional networks. One-dimensional convolutional networks could do classification problems for audio or biomedical sensor data. On the other hand, mostly-convolutional models like object detection and signal detection are a natural fit for extending the sophistication of our models without greatly changing the training algorithm. Really, many models that are mostly convolutional networks could be adapted to a thermodynamic system. Image segmentation and diffusion-based image generation are also possible directions. However, more sophisticated application of thermodynamic models may require more sophisticated host data processing. The fact that only binary data is passed between our blocks seriously interferes with the performance of a diffusion model, for instance. On the other hand, adding lightweight host processing or residual data streams between the primary thermodynamic blocks could open up a whole family of powerful thermodynamic models trained with simple variations or extensions of the framework presented here. 

\section{Acknowledgments}

A. G. M. would like to thank Professor Thomas Chen of the Department of Mathematics at the University of Texas at Austin for his supervision and guidance. He also thanks Professor Adam Klivans and the National Science Foundation through the Institute for Foundations of Machine Learning (IFML) for their support. 

\printbibliography

\appendix

\section{Pretraining}
\label{sec:pretraining}

The purpose of this sections is to provide supplemental algorithmic detail explaining how models are prepared for Gibbs regularization (\Cref{sec:regularize}). In short, we train a standard model with bounded activations (Phase 1), then convert it to the right shape (Phase 2), then train it as an STE across block boundaries (Phase 3). This gets the system close enough to an Ising-compatible solution that our main training algorithm becomes effective. We refer to this process as `pretraining' because it produces a trained model as input for our main training algorithm. 

We begin with Phase 1, training a conventional residual convolutional model, the WideResNet, whose block structure is conveniently
thermodynamically friendly. Every block contains an internal activation after the first convolution and an
output activation after the residual sum. Both are implemented with the penalized tanh activation \cite{xu2016revisesaturatedactivationfunctions}
\[
\mathrm{ptanh}(x)=
\begin{cases}
\tanh(x), & x\ge 0, \\
0.25\,\tanh(x), & x<0.
\end{cases}
\]
The sharp corner in the activation function provides a ReLU-like decision boundary which helps the network learn, while keeping overall activations bounded in a roughly sigmoid shape. 

The role of this phase is simply to learn a strong bounded-activation
teacher model before any binarization or Gibbs-stability constraints are imposed. Because every inter-block activation is already bounded, the model starts in a regime that is
structurally close to the binary communication pattern required by thermodynamic hardware, while still
being easy to optimize with ordinary deep-learning tools. 

In the implementation
we train this model with standard large-scale image-classification practice: SGD with Nesterov
momentum, cosine decay with warmup, label smoothing, dropout in the hidden path, and dataset-appropriate
augmentation including random crop, horizontal flips where appropriate, AutoAugment, and Cutout.

\subsection{Adaptation \& Folding}

Phase 2 of pretraining converts the bounded continuous teacher into an exact tanh-plus-affine parameterization that is
compatible with later binarization and Gibbs interpretation. We first replace every penalized tanh by a
fixed affine-transformed tanh approximation
\[
f(x)=a\tanh(bx+c)+d
\]
with
\[
a=0.625,\qquad b=1.239,\qquad c=-0.523,\qquad d=0.375.
\]
These constants were chosen to minimize $L^2$ loss against ptanh. We continue the training process until we recover the accuracy lost by the activation function replacement.

Finally, we perform an exact algebraic fold of the inner affine parameters $(b,c)$ into the preceding
BatchNorm layers. After this fold, each activation function has the form
\[
u \mapsto a \odot \tanh(u)+d.
\]
This fold is exact; it is not an approximation. We then move the $a$ and $d$ parameters of the output-layer activations to the input layer of the next block.  
This results in a network whose inter-block data is in the range $[-1,1]$ and is of the form $\tanh(\cdots)$. This network is now ready for binarization: by replacing the final tanh with sign, we produce a network based exclusively on affine transforms and tanh which passes only binary tensors with values in $\{\pm1\}$ between blocks. This is exactly the sort of raw material that we need for our Ising hardware implementation.

\subsection{Straight-Through Estimator}
\label{sec:ste}

Once the continuous pretraining is complete, we switch the architecture to a straight-through estimator (STE) and continue to train. The goal is to adapt the folded tanh network to binary inter-block communication. The binarization only happens between blocks: within a block, we still use tanh, because the block is conceptualized as a high-temperature mean-field Ising system. The block structure produced by phase 2 (after batch-norm folding) is
\begin{align}
    y(x) = \tanh(K_3 A_\text{in}x + K_2 A_\text{mid}\tanh(b + K_1A_\text{in}x))
\end{align}
We now replace the final activation function with $\zeta$, the surrogate sign function:
\begin{align}
    y(x) = \zeta(K_3 A_\text{in}x + K_2 A_\text{mid}\tanh(b + K_1A_\text{in}x))
\end{align}
During training, the boundary activation function $\zeta$ is a straight-through Bernoulli estimator of $\sign$. The
forward pass uses Bernoulli spins drawn from the corresponding local fields, while the backward pass uses
the derivative of $\tanh$ as a smooth surrogate. We use $8$ stochastic Bernoulli forward samples per step, adding noise tolerance and reducing gradient variance. 
During inference, on the other hand, $\zeta$ is set to the hard $\sign$ function. In other words,
\begin{align}
    \text{Inference: }\zeta(z) = \sign(z), && \text{Training: } \zeta(z) = \mathcal{B}(\sigma(2z)), && \frac{d \zeta}{dz} = 1-\tanh^2(z)
\end{align}
As usual for straight-through estimators, the forward and backward definitions of $\zeta$ are intentionally different: the forward pass uses a discrete stochastic estimator, while the backward pass uses the derivative of a smooth surrogate. Note that all inter-block data is now in $\{\pm1\}$, a critical part of our final model criteria. The internal activations are still continuous, and based on tanh. This is not an issue because the intra-block activations now represent mean-field averages of the fully binary Gibbs process. 

Rather than switching abruptly from continuous to
binary behavior, the script uses a short temperature-anneal period in which the boundary activation
interpolates between tanh and Bernoulli; after this warmup, training proceeds fully in Bernoulli/sign
mode. To reduce variance, the training step averages multiple stochastic forward samples of the same
minibatch.

The STE training also uses two teachers. The first is the original ptanh network, which provides a standard
logit-level distillation target. The second is a frozen copy of the tanh-adapted network,
which provides feature-level distillation on the outputs of the three residual stages. In addition, we
regularize the per-channel offsets toward zero in order to encourage centered binary activations, and we
penalize excessively large BatchNorm scales so that the subsequent folding step remains well behaved.
We train this STE as far as it will go before we start adding Gibbs regularization. 

Finally, we eliminate elements of the network which are not affine or tanh by a folding process. The parameters $b$ and $c$ are algebraically absorbed into the preceding batch normalization layer by the transformation $\text{BN}' = b \cdot \text{BN} + c$, and the batch normalization is
then folded into the convolution weights and biases.
After folding, each activation site has the form $a \odot \tanh(z) + d$, where
$z$ is the raw convolution output. 

\section{Algorithm Variations for Gibbs Regularization}

\subsection{Block-Weighted Magnitude Regularization}

It can sometimes help to adjust the target magnitudes of the magnitude regularizer according to block position. At block $b$ of $n_b$ total blocks, we can scale the target magnitude by 
    \begin{align}
        1+\frac{1}{5}\left(1-\frac{b}{n_b}\right).
    \end{align}
    This essentially tries to make earlier blocks more confident, which prevents the pile-up of stochastic error across blocks that can cause problems in deeper models. It should be noted that because the magnitude regularizer roughly determines the position of the main spike in the $\rho$ distribution, applying a block-weighted magnitude regularizer in this way will extend the period of fast convergence for earlier blocks. 

\subsection{Flip-Rate-Weighted Fixed-Point Loss}
\label{sec:fp_flipweighted}
In multi-block models, the per-block fixed-point losses
$\mathcal{L}_{\text{FP},i}$ of \Cref{sec:fp_loss} are aggregated to form a
single training signal. The default recipe used in our experiments simply
sums them,
\begin{equation}
    \mathcal{L}_\text{FP}^\text{total}
    \;=\; \sum_{i=1}^{n_b} \mathcal{L}_{\text{FP},i},
    \label{eq:fp_total_uniform}
\end{equation}
treating every block as equally important at every training step. A
natural alternative is to allocate gradient mass to each block in
proportion to how unstable that block currently is, so that whichever
block currently lags behind in Gibbs stability receives the largest share
of each update. We describe this variant here as a drop-in replacement for
\eqref{eq:fp_total_uniform}; on shallow models the two aggregations
behave essentially identically in practice, but on deeper or wider
networks one or two blocks can dominate the residual instability and
benefit from preferential updates.
We define the per-block flip rate $f_i \in [0, 1]$ as the fraction of spin
sites at block $i$ at which the Gibbs consensus $y_{\text{gibbs}}^{(i)}$
disagrees with the feed-forward STE sign of the block-output field
$z_2^{(i)}$:
\begin{equation}
    f_i \;=\; \left\langle \bm{1}\!\left[\,\sign z_2^{(i)}
    \,\ne\, y_{\text{gibbs}}^{(i)}\,\right] \right\rangle,
    \label{eq:flip_rate}
\end{equation}
where $\langle \cdot \rangle$ denotes the empirical mean over the batch
and over the spatial--channel positions of block $i$. The flip rate is
read off the Gibbs chain that already runs to produce the fixed-point
target of \Cref{sec:fp_loss}, and is treated as a scalar with no
gradient. It is large when the block's feed-forward sign and its Gibbs
equilibrium disagree, and shrinks toward zero as the fixed-point loss
drives them into agreement.
Given $\{f_i\}_{i=1}^{n_b}$, form unnormalised weights with a fixed
baseline $\alpha > 0$,
\begin{equation}
    \tilde{w}_i \;=\; \alpha + f_i,
\end{equation}
and renormalise so the weights sum to $n_b$, preserving the overall scale
of the unweighted aggregate \eqref{eq:fp_total_uniform}:
\begin{equation}
    w_i \;=\; \frac{n_b\, \tilde{w}_i}{\sum_{j=1}^{n_b} \tilde{w}_j}
        \;=\; \frac{n_b\, (\alpha + f_i)}{n_b\,\alpha + \sum_{j=1}^{n_b} f_j}.
    \label{eq:flip_weights}
\end{equation}
The flip-rate-weighted aggregate that replaces
\eqref{eq:fp_total_uniform} is then
\begin{equation}
    \mathcal{L}_\text{FP}^\text{total}
    \;=\; \sum_{i=1}^{n_b} w_i\, \mathcal{L}_{\text{FP},i}.
    \label{eq:fp_total_weighted}
\end{equation}
We use $\alpha = 0.2$ in all of our experiments with this variant.
The construction has three convenient properties. First, when all blocks
have the same flip rate, $f_i \equiv c$, the weights collapse to
$w_i = 1$ and \eqref{eq:fp_total_weighted} reduces exactly to the
unweighted aggregate \eqref{eq:fp_total_uniform}; only \emph{relative}
differences in flip rate redistribute mass between blocks. Second, the
baseline $\alpha$ lower-bounds each block's share: as $\alpha \to 0$, a
fully-stable block ($f_i = 0$) would receive $w_i = 0$ and stop
receiving fixed-point gradient at all, while as $\alpha \to \infty$ the
weights flatten back to uniform and we recover \eqref{eq:fp_total_uniform}.
Third, the maximum amplification any single block can receive is
bounded:
\begin{equation}
    \max_i w_i \;\le\; \frac{n_b\,(1 + \alpha)}{1 + \alpha + (n_b - 1)\,\alpha},
    \label{eq:flip_weights_amplification}
\end{equation}
attained when the worst block has $f_k = 1$ and the others have
$f = 0$. For $n_b = 3$ and $\alpha = 0.2$ this caps amplification at
$w_k \le 2.25$, with the corresponding stable blocks receiving
$w \approx 0.375$.
Mechanistically, $w_i$ acts as a per-block, per-step learning-rate
multiplier that concentrates gradient mass on whichever block currently
sits at the bottleneck of Gibbs stability. Once the fixed-point loss
drives that block's flip rate down, the weighting redistributes to the
next bottleneck. The unweighted aggregate \eqref{eq:fp_total_uniform} is
the limit $\alpha \to \infty$ of the same scheme, and the two
aggregations agree exactly whenever the per-block flip rates are equal,
which is empirically the regime our shallowest models operate in for most
of Gibbs regularization training.

\section{Integrated Autocorrelation Time}
\label{sec:autocorrelation}

The missing piece in our convergence analysis is an estimate of the integrated autocorrelation time for each output spin. Numerical experiments show that in our convolutional Ising blocks, the measured per-spin integrated autocorrelation time is typically within $[1, 1.5]$---essentially, our chains are nearly i.i.d. We would like to justify, understand, or predict this theoretically. 

\subsection{Dobrushin Approximation}

To demonstrate that we can at least control the autocorrelation time,
it is sufficient to find an upper bound on the maximum integrated autocorrelation time $\tau_\text{int}^*$. We will start with the easy case: Glauber (not red-black Gibbs) dynamics and $\delta \rightarrow 0$. Consider the natural approximation of the Dobrushin constant
\begin{align}
    \tilde{\alpha}(\delta) := \max_{i} \sum_{j \neq i} \tanh(\delta |J_{ij}|).
\end{align}
Using $\tanh x \leq x$ for $x \geq 0$, define
\begin{align}
    \kappa := \max_i \sum_{j \neq i}|J_{ij}|.
\end{align}
Then $\tilde{\alpha}(\delta) \leq \delta\kappa$, and $\tilde{\alpha}(\delta)=\delta\kappa+\mathcal{O}(\delta^3)$ as $\delta\rightarrow0$ when the maximizing row is stable. A Dobrushin-type argument gives, for $\delta\kappa<1$, a bound of the form
\begin{align}
    \tau_\text{int}^* \leq \frac{C_0}{1-\tilde{\alpha}(\delta)} \leq \frac{C_0}{1-\delta\kappa},
\end{align}
where $C_0$ is independent of $\delta$ in this regime.
Combining this estimate with the leading stochastic-error scaling gives
\begin{align}
    R(G) \lesssim \frac{G^{-1/2}}{\delta\sqrt{1-\delta\kappa}}.
\end{align}
Within this loose bound, we can optimize the choice of $\delta$:
\begin{align}
    -\frac{1}{\delta}+\frac{\kappa}{2(1-\delta\kappa)}=0
    \quad\Longrightarrow\quad
    \delta = \frac{2}{3\kappa}.
\end{align}

This analysis has a major and unavoidable limitation: $\delta$ controls not only the convergence speed, but also the correctness of the final distribution $\mu$. For finitely many nonzero outputs, we know that there exists some $\Delta$ such that $0<\delta<\Delta$ implies that the signs of $\mu$ agree with the feed-forward signs. However, in practice, increasing $\delta$ comes with an accuracy tradeoff because the stable distribution drifts away from the low-$\delta$ `correct' distribution. In other words, the optimal $\delta$ for convergence rate to the stable distribution may not be the optimal $\delta$ for achieving accurate inference.  

However, in the context of the above estimate, this caveat is not actually very relevant: the fact is that the Dobrushin bound is simply far too loose to be useful, and the $\delta$ it predicts is so small that accuracy distortions aren't relevant in that regime. In the case of our CIFAR-10 network, the measured $\tilde{\alpha}(1)$ values of the blocks are in the region of 50 to 250, and we observe that optimal convergence speeds are occurring for $\delta$ larger than $1/10$. 

\subsection{Linearized Red-Black Mode Theory}

The Dobrushin analysis shows that autocorrelation does not explode as $\delta \rightarrow 0$. However, it is not a particularly useful bound, and only applies to $\delta$ small enough that it's no longer practically relevant. Now, we develop a more refined theory which is capable of giving us much tighter estimates, at the cost of only being an approximation. We will also present numerical experiments showing that, in practice, it is accurate and slightly conservative. 

The approach is to linearize the red-black Gibbs sampling, giving us a linearized update rule for each layer. We expand around the Jacobian of naive mean-field theory. This gives us a symmetric positive semidefinite operator. We then use eigendecomposition to determine the dominant modes, i.e. those with the largest eigenvalues. The resulting eigenvalues directly control the autocorrelation times of those modes. Our goal is to use this mode decomposition to obtain a practical upper bound on the per-spin integrated autocorrelation time. Since each coordinate observable is a convex combination of the modal observables, the top eigenvalue yields an immediate upper bound on all per-spin integrated autocorrelation times.

Consider a single thermodynamic block with two spin layers, denoted \(s_1\) and \(s_2\), updated by alternating red-black Gibbs sweeps. Let $K$ be the convolution coupling between the two layers, and 
\[
A := \operatorname{diag}(a_{\mathrm{mid}}).
\]
The conditional mean maps have the form
\begin{align}
\mathbb E[s_1^+ \mid s_2]
&=
\tanh\!\big(b_1 + \,\delta\, A K^\top s_2\big), \\
\mathbb E[s_2^+ \mid s_1]
&=
\tanh\!\big(\delta\,(K(A s_1 + d_{\mathrm{mid}})+b_2)\big),
\end{align}
where $b_1$ and $b_2$ collect the image-dependent biases given by the coupling with the input layer and the fixed bias, and $\delta$ is the post-training temperature scale.

Let $m_1,m_2$ denote the stationary means of the two layers, and define the diagonal susceptibility matrices
\begin{align}
D_1 := \operatorname{diag}(1-m_1^2), \qquad
D_2 := \operatorname{diag}(1-m_2^2).
\end{align}
These are the coordinatewise derivatives of the conditional mean maps at stationarity. We then introduce normalized centered fluctuations
\begin{align}
z_1 := D_1^{-1/2}(s_1-m_1), \qquad
z_2 := D_2^{-1/2}(s_2-m_2).
\end{align}
Symmetrically linearizing the two half-steps around stationarity gives
\begin{align}
z_1^+ &\approx H_{1\leftarrow 2} z_2,
&
H_{1\leftarrow 2}
&=\delta\, D_1^{1/2} A K^\top D_2^{1/2},
\\
z_2^+ &\approx H_{2\leftarrow 1} z_1,
&
H_{2\leftarrow 1}
&=
\delta\, D_2^{1/2} K A D_1^{1/2}.
\end{align}
Composing the two half-steps yields the same-layer full-sweep linearizations
\begin{align}
z_1^{(t+1)} &\approx S_1 z_1^{(t)}, &
S_1 &:= H_{1\leftarrow 2} H_{2\leftarrow 1},
\\
z_2^{(t+1)} &\approx S_2 z_2^{(t)}, &
S_2 &:= H_{2\leftarrow 1} H_{1\leftarrow 2}.
\end{align}
Explicitly,
\begin{align}
S_1
&=
\,\delta^2\,
D_1^{1/2} A K^\top D_2 K A D_1^{1/2},
\\
S_2
&=
\,\delta^2\,
D_2^{1/2} K A D_1 A K^\top D_2^{1/2}.
\end{align}
These operators are symmetric positive semidefinite. Indeed, if we define
\begin{align}
L := \;\delta\; D_2^{1/2} K A D_1^{1/2},
\end{align}
then
\begin{align}
S_1 = L^\top L, \qquad S_2 = L L^\top.
\end{align}
Now, let \((\lambda_r,u_r)\) denote an orthonormal eigensystem of \(S_2\):
\begin{align}
S_2 u_r = \lambda_r u_r
\end{align}
Suppose $\delta$ is small enough that $0 \le \lambda_r < 1$.
The nonzero eigenvalues of \(S_1\) and \(S_2\) agree. For the modal observable
\begin{align}
a_r(t) := \langle u_r, z_2^{(t)} \rangle,
\end{align}
we have the approximate relation
\begin{align}
\mathbb E[a_r(t+1)\mid a_r(t)] \approx \lambda_r a_r(t),
\end{align}
so the modal lag-\(g\) autocorrelation is roughly
\begin{align}
\rho_r(g) \approx \lambda_r^g.
\end{align}
Therefore the infinite integrated autocorrelation time of mode \(r\) is
\begin{align}
\tau_r
=
1+2\sum_{g=1}^{\infty}\lambda_r^g
=
\frac{1+\lambda_r}{1-\lambda_r}.
\end{align}
Thus the eigenvalues have a direct dynamical meaning: small \(\lambda_r\) correspond to nearly i.i.d.\ modes, while eigenvalues close to \(1\) correspond to persistent slow modes.

A useful metric of the spinwise concentration of a mode is the participation ratio
\begin{align}
\mathrm{PR}(u_r)=\frac{1}{\sum_i u_r(i)^4},
\end{align}
which distinguishes localized modes from broad collective modes. Broad slow modes are precisely the dangerous case for correlated many-spin errors, while localized slow modes affect only a small subset of coordinates.

\subsubsection{Dependence on \(\delta\)}
The operator depends on \(\delta\) both explicitly and through the stationary susceptibilities \(D_1(\delta),D_2(\delta)\). Near \(\delta=0\), however, the structure is especially simple. Since the second bank is driven by a field scaled by \(\delta\), we have \(m_2(\delta)=O(\delta)\), hence \(D_2(\delta)=I+O(\delta^2)\). The first bank has a nonzero forward field already at \(\delta=0\), so \(m_1(\delta)=m_1(0)+O(\delta^2)\), hence \(D_1(\delta)=D_1(0)+O(\delta^2)\). It follows that
\begin{align}
S_1(\delta) &= \delta^2 \,\bar S_1 + O(\delta^4), \\
S_2(\delta) &= \delta^2 \,\bar S_2 + O(\delta^4),
\end{align}
for appropriate \(\delta\)-independent limiting operators \(\bar S_1,\bar S_2\). Consequently each eigenvalue satisfies
\begin{align}
\lambda_r(\delta)=\delta^2 \bar\lambda_r + O(\delta^4),
\end{align}
and hence
\begin{align}
\tau_r(\delta)
=
\frac{1+\lambda_r(\delta)}{1-\lambda_r(\delta)}
=
1 + 2\bar\lambda_r \delta^2 + O(\delta^4).
\end{align}
Thus the modal integrated autocorrelation times remain close to \(1\) as \(\delta\to 0\). In particular, there is no blow-up of autocorrelation time in the low-\(\delta\) regime.

\begin{figure}[h]
    \centering
    \includegraphics[width=0.6\linewidth]{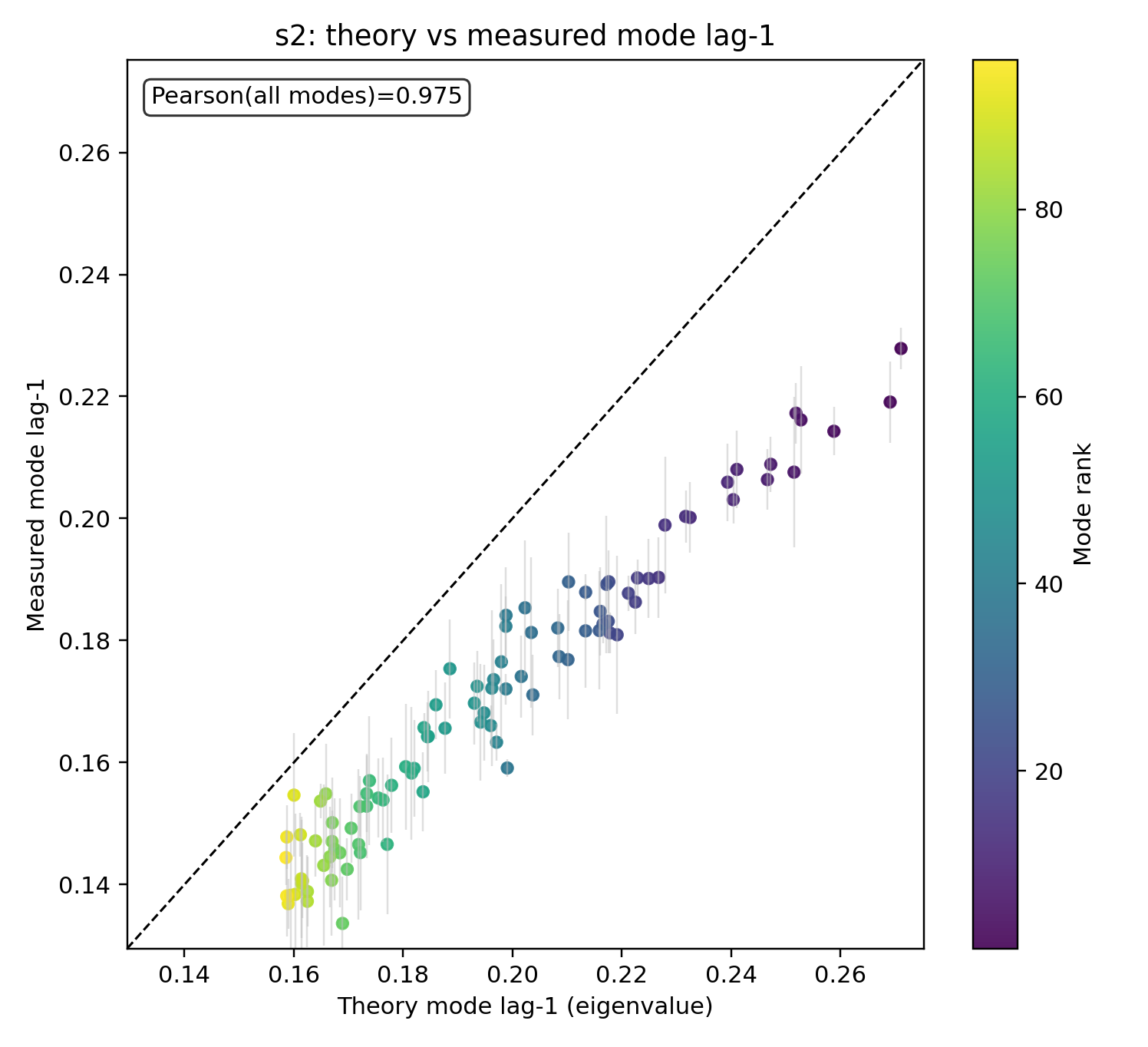}
    \caption{Predicted vs. measured lag-1 autocorrelation function for dominant eigenmodes of the $L^\top L$ and $LL^\top$ operators.}
    \label{fig:modes}
\end{figure}

\subsubsection{Experimental Validation on Modal Observables}
The theory above was tested on its natural observables, namely the eigenmodes themselves. For each retained mode \(u_r\), the empirical red-black Gibbs trajectory was projected onto \(u_r\), and the resulting scalar mode process was used to estimate the lag-1 autocorrelation. When compared against the theoretical prediction \(\lambda_r\), the agreement was very strong: across the top \(96\) modes, the Pearson correlation between predicted and measured lag-1 modal autocorrelation was approximately \(0.980\) for bank \(s_1\) and approximately \(0.975\) for bank \(s_2\). Weighted correlations were slightly higher still. This is strong evidence that the \(L^\top L\) theory captures the persistent same-bank modal dynamics of full red-black sweeps with high accuracy.

It is important to emphasize that modal validation is the correct comparison for this theory. Coordinatewise truncated integrated autocorrelation can be affected by short-lag oscillatory effects and by the truncation of the empirical sum, whereas the present PSD theory is intentionally a theory of the persistent same-bank envelope. For the purpose of obtaining upper bounds on infinite integrated autocorrelation times, however, this is the correct level of description.

\subsubsection{Per-Spin Integrated Autocorrelation from Modes}
Within the same linearized PSD model, the integrated autocorrelation time of a coordinate is a convex combination of the modal values. If \(u_r(i)\) denotes the \(i\)th coordinate of the normalized eigenvector \(u_r\), then
\begin{align}
\tau_i
=
\sum_r u_r(i)^2 \frac{1+\lambda_r}{1-\lambda_r}.
\end{align}
This immediately implies the global bound
\begin{align}
\tau_i \le \frac{1+\lambda_1}{1-\lambda_1},
\end{align}
where \(\lambda_1\) is the top eigenvalue. Hence the maximum per-spin integrated autocorrelation time satisfies
\begin{align}
\tau_{\mathrm{int}}^* \le \frac{1+\lambda_1}{1-\lambda_1}.
\end{align}
Since \(\lambda_1(\delta)=O(\delta^2)\), we conclude that
\begin{align}
\tau_{\mathrm{int}}^*(\delta)=1+O(\delta^2).
\end{align}
This is the central theoretical conclusion needed for the convergence analysis: in the low-\(\delta\) regime, the difficulty of sign recovery is controlled primarily by the smallness of the means \(\mu_i\), not by any growth in the integrated autocorrelation time.

A sharper coordinatewise upper bound is available from the top \(k\) modes. Define
\begin{align}
\phi(\lambda):=\frac{1+\lambda}{1-\lambda}.
\end{align}
Then monotonicity of \(\phi\) gives
\begin{align}
\tau_i
\le
\sum_{r=1}^{k} u_r(i)^2 \phi(\lambda_r)
+
\Big(1-\sum_{r=1}^{k} u_r(i)^2\Big)\phi(\lambda_k).
\end{align}
Thus the top modes do not merely explain the large-scale dynamics qualitatively; they provide explicit rigorous upper bounds on every per-spin integrated autocorrelation time within the linearized theory.

\subsubsection{Consequences for Convergence of Time Averages}
The purpose of introducing \(\tau_{i,\mathrm{int}}\) is to control the effective sample size of the time average of spin \(i\). If
\begin{align}
m_i^{(n)} := \frac1n \sum_{t=1}^n \sigma_i^{(t)},
\end{align}
then under the convention
\begin{align}
\tau_{i,\mathrm{int}} = 1 + 2\sum_{g=1}^\infty \rho_i(g),
\end{align}
the asymptotic variance of the time average is
\begin{align}
\operatorname{Var}(m_i^{(n)})
\approx
\frac{(1-\mu_i^2)\tau_{i,\mathrm{int}}}{n}.
\end{align}
Therefore the number of sweeps required to determine the correct sign of \(m_i^{(n)}\) with high confidence is controlled by
\begin{align}
n \gtrsim
\frac{1-\mu_i^2}{\mu_i^2}\,
(\text{confidence factor})\,
\tau_{i,\mathrm{int}}.
\end{align}
Using the modal upper bound yields the sufficient condition
\begin{align}
n \gtrsim
\frac{1-\mu_i^2}{\mu_i^2}\,
(\text{confidence factor})\,
\frac{1+\lambda_1}{1-\lambda_1}.
\end{align}
In particular, since \(\lambda_1(\delta)=O(\delta^2)\), the autocorrelation correction stays bounded and close to \(1\) as \(\delta\to 0\). The dominant low-\(\delta\) scaling therefore comes from the factor \(\mu_i^{-2}\), not from any growth of autocorrelation.

\subsubsection{Summary}
The red-black full-sweep dynamics admits a natural linearization in normalized coordinates whose same-bank persistent dynamics is governed by the symmetric PSD operators \(S_1=L^\top L\) and \(S_2=LL^\top\). Their eigenvalues determine modal autocorrelation times through
\begin{align}
\tau_r=\frac{1+\lambda_r}{1-\lambda_r},
\end{align}
and each per-spin integrated autocorrelation time is a convex combination of these modal values. Hence
\begin{align}
\tau_i \le \frac{1+\lambda_1}{1-\lambda_1},
\end{align}
with the sharper top-\(k\) refinement above. Empirically, as seen in \Cref{fig:modes}, the modal lag-1 autocorrelations are predicted with Pearson correlation approximately \(0.98\), validating the theory at the level of its natural observables. Since \(\lambda_1(\delta)=O(\delta^2)\), the resulting autocorrelation bound remains close to \(1\) in the low-\(\delta\) regime.

\end{document}